\pdfoutput=1

\documentclass[a4paper,twoside]{article}

\usepackage{apalike}

\usepackage{amsmath}
\usepackage{amssymb}
\usepackage{mathtools}
\usepackage{epsfig}
\usepackage{tikz}
\usepackage{pgfplots}
\pgfplotsset{compat=1.18}

\usepackage{booktabs}
\usepackage{tabularx}
\usepackage{graphicx}
\usepackage[T1]{fontenc}
\usepackage[utf8]{inputenc}
\usepackage{array}
\usepackage{makecell}
\usepackage{caption}
\usepackage{url}
\newcolumntype{C}[1]{>{\centering\arraybackslash}p{#1}}
\usepackage[hidelinks]{hyperref}
\usepackage{lmodern}
\usepackage{array,booktabs,adjustbox}
\usepackage{array} 
\usepackage{multirow}
\usepackage{booktabs}
\usepackage{amsmath}
\usepackage{placeins}
\usepackage{subcaption}
\usepackage{SCITEPRESS}     

\begin{document}

\title{Sliced-Wasserstein Distribution Alignment Loss Improves the Ultra-Low-Bit Quantization of Large Language Models}

\author{
\authorname{
Deyu Cao\sup{1}\orcidAuthor{0009-0003-8563-3842}\thanks{These authors contributed equally to this work.},
Yixin Yin\sup{2}\orcidAuthor{0009-0009-7488-4754}\footnotemark[2],
and Samin Aref\sup{3}\orcidAuthor{0000-0002-5870-9253}
}
\affiliation{\sup{1}Department of Information and Communication Engineering, The University of Tokyo, 7-3-1 Hongo, Bunkyo-ku, Tokyo, Japan}
\affiliation{\sup{2}Department of Computer Science, University of Toronto, 40 St. George St, Toronto, Canada}
\affiliation{\sup{3}Department of Mechanical and Industrial Engineering, University of Toronto, 5 King's College Rd, Toronto, Canada}
\email{yixin.yin@mail.utoronto.ca, sou-tokuyu888@g.ecc.u-tokyo.ac.jp, s.aref@utoronto.ca}
}

\abstract{
The benefits of most large language models come with steep and often hidden economic and environmental costs due to their resource usage inefficiency during deployment. Model quantization improves energy and memory efficiency through representing model parameters by lower-precision values. However, compression below 4-bits often distorts activation distributions and degrades performance. We address this challenge by introducing a sliced Wasserstein loss function for distribution-aware calibration in ultra-low-bit post-training quantization. The proposed loss aligns the output distributions of full-precision and quantized models under random linear projections, complementing standard mean-squared error loss without adding any computational overhead during inference. Our proposed loss function can be incorporated with any post-training quantization framework that has a retraining component. We demonstrate the performance gains of our proposed model by incorporating it with two frontier methods known as OmniQuant and TesseraQ. Compared to these two baselines, the proposed loss consistently improves both perplexity and downstream task accuracy across multiple ultra-low-bit settings. Our proposed loss function recovers 4.12-20.37\% of the OmniQuant's lost accuracy on the language model LLaMA-2-7B, 0.93–7.65\% on OPT-6.7B, and 2.26–6.20\% on LLaMA-2-13B. TesseraQ's accuracy degradation is recovered by 3.63-7.63\% in relative terms when augmented by our proposed loss function. Taken together, these results demonstrate that distributional alignment provides a simple yet effective performance boost that can push the limits of frontier quantization methods. Our method is available on \href{https://github.com/TokuyuSou/SWLoss_LightCompress}{GitHub} to facilitate future progress in ultra-low-bit quantization.\\ \\
\textit{This is a post-peer-review accepted manuscript from the proceedings of the 18th International Conference on Agents and Artificial Intelligence (ICAART'26). The publisher authenticated version (version of record) and full citation details are available on the conference and SciTePress websites.}}


\maketitle

\clearpage

\section{Introduction}
Large Language Models (LLMs) have achieved remarkable capabilities, but their sheer size comes with a substantial computational and memory burden which translates to steep economic and environmental costs. Their high demands for electricity and water makes it challenging to deploy most high-precision LLMs in a resource efficient manner or in a low-resource environment. Quantization of LLMs has emerged as a central technique to alleviate these issues by representing model parameters in lower-precision numeric formats, drastically reducing their resource consumption~\cite{lang2024comprehensive} with minimal performance degradation.

Practically, 4-bit quantization has succeeded in compressing LLMs with hardly any loss in accuracy~\cite{jin2024comprehensive}, making it possible to fine-tune 65B-parameter models on a single GPU through 4-bit adapters. However, it has been proven difficult to go below 4-bit precision (ultra-low-bit) because extreme compression using the same methods adds a high quantization error~\cite{zhang2024efficientqat}. Naive ultra-low-bit quantization can misrepresent the knowledge learned by a model~\cite{liao2024apiq} and even result in severe model forgetting when fine-tuned downstream~\cite{dettmers2024qlora}.

But the allure of ultra-low-bit approaches is strong. For instance, a further 50\% reduction in memory usage can be achieved if a 4-bit model can be quantized to 2-bits without substantial performance degradation. The potential for maintaining fidelity while reducing resource consumption of LLMs  through ultra-low-bit quantization is substantial. Moreover, the advances in this area facilitate deploying or fine-tuning LLMs on edge devices and other hardware-restricted settings. An active research line has recently emerged for ultra-low-bit quantization techniques~\cite{cao2025saliency} that retain acceptable model performance while compressing parameters below 4 bits. 

In this study, we investigate the potential suitability of a distribution-matching loss function for improving frontier ultra-low-bit quantization methods. These types of loss functions are unique in the sense that they allow model training by comparing entire distributions of data, rather than individual data points. A shift from point-wise to distribution-wise measurement of training error has been successful in other areas of Artificial Intelligence (AI) including generative AI. However, to the best of our knowledge, they are yet to be explored for Post-Training Quantization (PTQ) of LLMs. In this study, we make the following contributions:

\begin{itemize}
    \item 
We provide the first investigation on suitability of distribution-matching loss functions for LLM quantization.
\item 
We experiment quantization with both Kullback-Leibler divergence loss and the sliced Wasserstein loss functions and identify the limitations and advantages of each these two novel approaches.
\item 
We design and implement a simple and effective usage of distribution-matching loss and demonstrate how it outperforms existing methods by offering an improved performance-precision tradeoff.
\item 
We augment two frontier quantization methods and test them on language models of different sizes and families.
\item 
We make our proposed method publicly available to facilitate future research.
\end{itemize}

This paper is structured as follows: Section \ref{s:literature} provides a review of the recently developed quantization methods. Section \ref{s:method} provides technical background and introduces the methodology. Section \ref{s:results} provides the experiment results based on two frontier methods as baselines: OmniQuant~\cite{shao2023omniquant} (in Section \ref{ss:OmniQuant}) and TesseraQ~\cite{li2024tesseraq} (in Section \ref{ss:TesseraQ}). Section \ref{sec:discuss} provides a discussion of the main findings and directions for future research. Additional results and details are provided as Supplementary Information (SI) at the end of this ArXiv manuscript.

\section{Literature Review}
\label{s:literature}

In this section, we review the current literature on ultra-low-bit quantization methods to situate our approach within the context of recent developments. For each study, we summarize the method and their key results. For a more detailed review of the literature, one may refer to~\cite{lang2024comprehensive,shen2024exploring,gong2025survey,tripathi2025survey}.

Jin et~al.\ present a structured framework
to evaluate quantization in instruction-tuned LLMs along three
dimensions: knowledge and capacity, alignment, and efficiency across ten benchmarks~\cite{jin2024comprehensive}.
Their experiments show the feasibility of nearly lossless 4-bit weight quantization in most tasks. They also show that LLMs with larger parameter count continue to outperform smaller ones even at the 4-bit compression level.  
A notable observation from~\cite{jin2024comprehensive} is that perplexity is correlated with the accuracy of downstream tasks for many quantized models. This suggests that reductions in perplexity can serve as a reliable proxy for preserving end-task performance. Building on this insight, they argue that most perplexity degradation often come from a small set of outlier weights that are poorly represented under aggressive quantization. To address this, they highlight outlier weight isolation as a promising path toward ultra-low-bit quantization. The SpQR technique~\cite{dettmers2024spqr} directly targets these outliers and thereby enables almost lossless 2-bit quantization after separating the outliers. By correcting for the dominant sources of quantization error, SpQR achieves results substantially better than the method GPTQ~\cite{frantar2022gptq} at the same bit width.

These findings suggest that with appropriate outlier-aware calibration, 2-bit models can approach the performance of their 16-bit counterparts. However, achieving this in practice remains challenging: mixed-precision quantized models often introduce hardware incompatibilities. As a result, while promising, deploying such models in real systems still demands considerable engineering effort. Motivated by this, Ahmed at~al. proposes a GPU-accelerated discrete optimization heuristic that quantizes LLMs without the need for outlier separation~\cite{ahmed2025discrete}. Their method is motivated by the critical perspective that in such mixed-precision models, up to 30\% of the parameters~\cite{ahmed2025discrete}, might be considered as outliers and left in full-precision format to limit the performance degradation. To produce fully quantized models that can be run on edge devices without FP16 arithmetic, they demonstrate the quantization of OPT-125m without outlier separation. They report 14\% lower perplexities on average for 3-bit quantization compared to four baseline methods without outlier separation: GPTQ~\cite{frantar2022gptq}, SqeezeLLM~\cite{kim2023squeezellm}, OmniQuant~\cite{shao2023omniquant}, and ApiQ~\cite{liao2024apiq}.

Lang et~al. conduct a large‑scale survey of post‑training quantization~(PTQ) and quantization‑aware training~(QAT) techniques for LLMs ~\cite{lang2024comprehensive}.  
They analyze the mathematical foundations of quantization and benchmark several widely used methods on standard LLM workloads.  
One of their key observations is that the effectiveness of quantization is both \emph{method- and precision-dependent}. For example, GPTQ~\cite{frantar2022gptq} has the best accuracy reported for the 4-bit level but deteriorates markedly at 3~bits, indicating that it cannot be naively scaled to lower bit-widths. In contrast, outlier-aware approaches such as SpQR~\cite{dettmers2024spqr} achieve competitive or superior performance in the 2-bit regime, highlighting that the optimal choice of method varies with the target precision.
Another example is the QAT method LLM-QAT~\cite{liu2023llm}, which adopts a mixed-precision configuration for weights and activations. Specifically, it uses 4-bit weights and key–value caches while keeping activations at 8-bit precision. This mixed-precision configuration outperforms in accuracy compared to applying 4-bit uniformly across all components.
This result supports the idea that performance gains can be achieved by allocating higher precision to more sensitive parameters or groups of parameters (e.g., activations or specific layers) at the cost of having a mixed-precision quantized model.
Lang et al.\ also report that EfficientQAT~\cite{zhang2024efficientqat} surpasses previous QAT approaches for models ranging from 7B to 70B parameters, indicating continued room for algorithmic improvement.  
Their study summarizes recent quantization strategies and trade-offs, demonstrating that 4-bit compression is now routinely achievable while highlighting opportunities to further compress models while maintaining performance.

Liao et~al. focus on the fine-tuning of LLMs under ultra-low-bit constraints and introduce the 2-bit quantized model framework ApiQ~\cite{liao2024apiq}. ApiQ seeks to recover the information lost due to extreme quantization through joint learning of low-rank adaptation (LoRA)~\cite{hu2022lora} modules along with quantization of model weights. In essence, the method preserves the original model's activations at the quantization point to prevent the knowledge distortion that arises with standard QLoRA~\cite{dettmers2024qlora} at 2-bit or lower precisions, which performs poorly due to quantization-induced forgetting. Through experiments on several language tasks and LLMs, ApiQ consistently outperforms baselines (including standard QLoRA) at 2-bit and 3-bit precisions. Notably, at 3-bit precision, ApiQ achieves fine-tuning performance that is comparable to full-precision (16-bit) fine-tuning, despite having much lower memory and computational costs. Even at 2-bit, where most existing approaches fail to produce usable quantized models, ApiQ remains robust and in some tasks closely matches full-precision performance. These results constitute a major breakthrough for ultra-low-bit fine-tuning. The success of ApiQ indicates the importance of mitigating quantization-induced forgetting and highlights methods that combine quantization with adaptation as effective solutions.

Cao and Aref propose enhancements to ultra-low-bit quantization by improving ApiQ's 2-bit performance through selective retraining~\cite{cao2025saliency}. They observe that partial updates alone do not recover the full representational capacity that full QAT offers, especially in data-scarce settings. Motivated by this, they introduce a saliency-aware partial retraining method. Rather than fine-tuning the full model or naively applying QAT, they regularize the partial fine-tuning process by incorporating a regularization term that encourages preservation of the most “salient” weights during quantization. They define saliency to be the relative importance of a weight to the model’s loss. Weights with larger gradients or higher sensitivity to perturbation are considered more salient, and are therefore prioritized for preservation. This strategy helps maintain essential parameters intact under 2-bit constraints. Their evaluations on models from the LLaMA family demonstrate that this saliency-aware method recovers the accuracy degradation of 2-bit ApiQ quantized models by 7.54\%-10.85\% in relative terms. Importantly, this improvement comes with minimal training overhead, as the model still avoids full retraining. Overall, their approach offers a practical technique for boosting performance in ultra-low-bit regimes.

Zhou et al.\ ~\cite{zhou2025lowra} introduce LowRA, a method designed to enable fine-tuning of large language models for average bit-widths below 2, i.e., model updates requiring less than 2 bits per parameter on average. Although techniques like QLoRA~\cite{dettmers2024qlora} have made efficient 4-bit adapters, scaling LoRA to 2-bit (and below) has been an open question that is investigated in~\cite{zhou2025lowra}. LowRA addresses this challenge by tuning a high-granularity quantization scheme for LoRA updates. It learns optimal quantization mappings and thresholds for each weight, and assigns non-uniform bit widths to different parameters. To maintain efficiency, it leverages custom CUDA kernels that support fine-tuning computation that is high-throughput and bit-flexible.
This approach enables LowRA to achieve extreme compression with minimal performance loss. In evaluations on four LLMs across multiple benchmarks, LowRA achieves strong accuracy not only at precisions above 2 bits but also at extremely low effective precision, down to 1.15 bits per parameter. Fine-tuning memory usage is reduced by up to 50 percent from baseline 2-bit methods. Their findings demonstrate the practical feasibility of sub-2-bit LoRA adaptation, enabling scalable and accurate LLM fine-tuning even on resource-constrained hardware.

In summary, the literature shows that 4-bit quantization is now a well-established technique that preserves LLM performance, and the current frontier is to advance models for the lower bit-width regimes. 
To address the accuracy degradation caused in this ultra-low-bit regime, researchers have proposed a variety of solutions. Outlier-aware quantization~\cite{frantar2022gptq} is an algorithmic solution that targets problematic weight outliers. Mixed-precision allocation~\cite{zhang2024efficientqat} is another algorithmic solution which assigns different bit-widths to different components of the model. Other methods are training-based. For example, ApiQ~\cite{liao2024apiq} and LowRA~\cite{zhou2025lowra} apply LoRA-augmented fine-tuning. Another approach is saliency-based partial updates~\cite{cao2025saliency}, which selectively fine-tune only the most important parameters. The common theme is that some form of compensation or adjustment needs to be applied when quantizing to 2-bit in order to preserve the learned knowledge of the model. This can be a mild reweighting of the model's weights~\cite{frantar2022gptq}, the inclusion of learned adapters~\cite{dettmers2024qlora}, or heuristic calibrations~\cite{dettmers2022calibration}. Another key consideration is the need to decide what piece of information is the most valuable to be preserved at high precision to reduce performance loss.

Finally, the reviewed literature emphasizes that algorithmic innovations alone are often insufficient to realize the full benefits of ultra-low bit quantization \cite{liao2024apiq,zhou2025lowra}. System-level support, such as efficient low-bit kernels \cite{zhou2025lowra} or hardware-aware implementations \cite{zhang2024efficientqat}, is often necessary to ensure actual speedups and memory savings. Highly quantized LLMs can perform competitively with their full-precision counterparts when these these considerations are addressed in the design of quantization methods, possibly through novel approaches. Despite the success of outlier-aware methods~\cite{frantar2022gptq,dettmers2024spqr}, mixed-precision and QAT approaches~\cite{liu2023llm,zhang2024efficientqat}, and LoRA-based fine-tuning frameworks ~\cite{liao2024apiq,zhou2025lowra}, the idea of employing distribution-matching loss functions for quantization has been underexplored. Existing approaches typically use simple objectives—mean squared error (MSE) for intermediate representations and KL divergence-based losses for final outputs—leaving the design of more principled loss functions largely unexplored. We aim to address this gap by investigating the sliced Wasserstein loss in LLM quantization. This simple and effective loss function has been widely adopted for (i) generative modeling~\cite{deshpande2019max,kolouri2018sliced} and (ii) distribution alignment in unsupervised domain adaptation~\cite{lee2019sliced}.

\section{Methodology}
\label{s:method}

\subsection{Evaluation measures}
\label{sec:measurement}

To understand how quantization affects large language models beyond the common measure of perplexity, we adopt a set of complementary metrics that capture both modeling fidelity and behavioral robustness. Although perplexity in benchmarks such as WikiText-2~\cite{merity2016wikitext} and C4~\cite{raffel2020exploring} remain our principal indicator of language modeling quality, we argue that perplexity alone cannot fully reflect subtle distortions in activation distributions and downstream reasoning.

Hence, we extend our evaluation along two axes. (1) Downstream task accuracy is assessed through zero-shot evaluation on six common-sense reasoning benchmarks using the LM evaluation harness framework~\cite{eval-harness}: ARC-Challenge and ARC-Easy~\cite{clark2018think}, BoolQ~\cite{clark2019boolq}, HellaSwag~\cite{zellers2019hellaswag}, PIQA~\cite{bisk2020piqa}, and WinoGrande~\cite{sakaguchi2021winogrande}. These tasks allow us to trace how local quantization errors propagate to higher-level reasoning capabilities. (2) Training efficiency metrics—including the memory footprint of the GPU and the calibration training time—are analyzed (and reported in Table \ref{tab:compute_llama_7b_13b}) to characterize the computational overhead of our proposed method. Since the sliced-Wasserstein loss is applied only during the calibration phase and introduces no inference-time cost, we do not measure inference speed or memory usage, which remain unchanged from the baseline method.

Together, these metrics expose an important observation: even when perplexity remains low, quantized models may drift in distributional behavior, showing miscalibration or degraded consistency. This discrepancy motivates us to take a deeper look at the loss design itself. Instead of matching outputs point by point through MSE alone, we seek to align their distributions, a direction we explore further through the Sliced-Wasserstein Loss introduced in Section \ref{s:sw-distance}.

\subsection{Kullback–Leibler divergence calibration}
\label{s:kl}
A natural distributional objective for calibration is the Kullback--Leibler (KL) divergence between the token distributions of the full-precision and quantized models. For an input $x$, let $p(\cdot\mid x)$ and $q(\cdot\mid x)$ denote the corresponding output distributions. The KL divergence objective minimizes 
\[
\mathcal{L}_{\mathrm{KL}}(p\parallel q)
= \sum_{i} p_i(x)\,\log\frac{p_i(x)}{q_i(x)}.
\]
This encourages the quantized model to preserve relative token likelihoods and, in principle, aligns calibration with language modeling objectives more directly than it does under a pure activation MSE objective.

In practice, when we integrated a KL term into the \emph{OmniQuant} calibration loop, the gains were modest and inconsistent across bit-widths and datasets. We observed sensitivity to temperature/label smoothing and to the calibration set composition. This sensitivity was such that small changes could swing results from slight improvements to regressions. Moreover, KL focuses on the final softmax layer and provides a pointwise distribution match per token, offering limited leverage over the broader geometry of hidden activations where ultra-low-bit artifacts often originate.

These observations motivated our shift toward a \emph{distribution matching at the representation level}: rather than only aligning output probabilities, we aim to align the \emph{shapes} of the block outputs themselves. Sections \ref{s:sw-distance}--\ref{subsec:method_sw} introduce a Sliced-Wasserstein (SW) loss that complements MSE by enforcing distributional alignment under random projections, which can potentially yield more stable and consistent improvements in the ultra-low-bit regime.

\subsection{The sliced‐Wasserstein distance and loss function}
\label{s:sw-distance}

\label{subsec:sw_background}

The Sliced-Wasserstein (SW) distance is a computationally advantageous approximation~\cite{cuturi2013sinkhorn} of the Wasserstein distance between two distributions~\cite{tanguy2025Wasserstein}. Unlike the Wasserstein distance, the SW distance operates in a one-dimensional space by averaging the distances obtained from multiple random slices (projections) of both distributions. The SW loss function compares a pair of high-dimensional predicted and target distributions by integrating one-dimensional SW distances. This distribution-matching loss function is differentiable and therefore its gradient can flow back for updating model parameters though a numerical optimization-based training process~\cite{tanguy2025Wasserstein}. Moreover, it has been effectively utilized in various deep learning contexts—such as generative modeling and domain adaptation—as a computationally efficient means of aligning high-dimensional feature distributions~\cite{deshpande2019max,kolouri2018sliced,lee2019sliced}.

\subsection{Sliced‐Wasserstein loss for quantization}
\label{subsec:method_sw}
Our proposed method is general and can be incorporated with QAT methods (especially those that seek to preserve activations of intermediate layers) as well as PTQ methods that involve some retraining. However, we focus our methodological explanations on how the proposed SW loss function can be incorporated into a PTQ pipeline using the OmniQuant method ~\cite{shao2023omniquant} as an example. Our experiments and discussions will be based on two frontier quantization methods: OmniQuant~\cite{shao2023omniquant} and TesseraQ~\cite{li2024tesseraq}. 

The foundation of the OmniQuant method formulates quantization of each transformer block via block‐wise error minimization as in \eqref{eq:omni_block_loss}.

\begin{equation}
\begin{split}
\min_{\Theta_1,\Theta_2}\; \mathbb{E}_{W,X}\;\big\|F(W,X)\;-
\;F(Q_w(W;\Theta_1),\,
\\
Q_a(X;\Theta_2))\big\|_{2}^{2}\;
\label{eq:omni_block_loss}
\end{split}
\end{equation}

In Eq. \eqref{eq:omni_block_loss}, \(W\) and \(X\) denote the full‐precision weights and activations of a block. \(Q_w\) and \(Q_a\) are the quantizers parameterized by learnable quantization parameters \(\Theta_1\) and \(\Theta_2\)  respectively. \(\Theta_1\) denotes the clipping thresholds in learnable weight clipping. \(\Theta_2\) denotes equivalent transformations in learnable equivalent transformation. 

We build upon this foundation to augment the block-wise optimization with a distributional term based on the SW distance~\cite{bonneel2015sliced} between the full-precision block output \(Y_{\mathrm{fp}}\) and the quantized block output \(Y_{\mathrm{q}}\).

Consider that each output has the shape \([\text{batch}, \text{seq\_len}, d]\). We reshape them into \(N\) samples in \(\mathbb{R}^{d}\), where \(N = \text{batch} \times \text{seq\_len}\) is the number of samples and \(d \) is the dimension of each sample. Specifically, we flatten the batch and sequence dimensions to obtain matrices \(Y_{\mathrm{fp}}, Y_{\mathrm{q}} \in \mathbb{R}^{N \times d}\) defined as follows.
$$
Y_{\mathrm{fp}} = F(W,X),
Y_{\mathrm{q}} = F\!\bigl(Q_w(W;\Theta_1),\,Q_a(X;\Theta_2)\bigr)
\label{eq:fp_q_def}
$$
Computing the exact Wasserstein distance in high-dimensional space is intractable, since it requires solving a complex optimal transport problem with computational cost growing super-linearly with \(d\)~\cite{cuturi2013sinkhorn}. However, for one-dimensional distributions, the Wasserstein-1 distance has a simple closed-form expression based on sorting~\cite{bonneel2015sliced}. To leverage this tractability, we use the SW distance which approximates the high-dimensional distance by averaging multiple 1D Wasserstein distances over random linear projections (slices).

Specifically, we draw $n_{\text{proj}}$ random unit vectors $\{u_i\}_{i=1}^{n_{\text{proj}}} \subset \mathbb{S}^{d-1}$ (with $u_i \in \mathbb{R}^d$) and compute the projected values using the following two matrix multiplications:
\begin{equation}
p_{\mathrm{fp}}^{(i)} = Y_{\mathrm{fp}} u_i \in \mathbb{R}^{N},\qquad
p_{\mathrm{q}}^{(i)} = Y_{\mathrm{q}} u_i \in \mathbb{R}^{N}.
\label{eq:projection}
\end{equation}
Each projection yields \(N\) scalar values. After sorting each set of projected values, we define the empirical 1D Wasserstein-1 distance as
\begin{equation}
W_1^{(i)} \;=\; \frac{1}{N}\sum_{j=1}^{N}\!\left|\,\tilde p_{\mathrm{fp},j}^{(i)}-\tilde p_{\mathrm{q},j}^{(i)}\right|,
\label{eq:wasserstein_1d}
\end{equation}
where \(\tilde p_{\mathrm{fp},j}^{(i)}\) and \(\tilde p_{\mathrm{q},j}^{(i)}\) denote the \(j\)-th elements of the sorted projections. The sliced-Wasserstein loss is then represented as follows:
\begin{equation}
\mathcal{L}_{\mathrm{SW}} \;=\; \frac{1}{n_{\text{proj}}}\sum_{i=1}^{n_{\text{proj}}} W_1^{(i)}.
\label{eq:sliced_wasserstein_fixed}
\end{equation}

Finally, the overall loss for the block is obtained from a linear combination of the original mean‐squared error (MSE) loss \(\mathcal L_{\rm MSE} = \|Y_{\rm fp}-Y_{\rm q}\|_2^2\) and the SW loss using the weight $\textbf{sw\_w}\in[0,1]$:
\begin{equation}
\mathcal L_{\rm block} = (1-\textbf{sw\_w})\,\mathcal L_{\rm MSE} \;+\; \textbf{sw\_w}\,\mathcal L_{\rm SW}.
\label{eq:combined_loss}
\end{equation}

Intuitively, while the standard MSE term enforces point‐wise matching between full‐precision and quantized block outputs, the sliced‐Wasserstein term further ensures the \emph{distributional alignment} of the two outputs under multiple random linear projections. By doing so, we capture not only mean and variance discrepancies, but also differences among the shapes of the two distributions, thereby helping to account for unnatural quantization artifacts and improve downstream fidelity.

We integrate this loss into the layer-wise sequential optimization pipeline of OmniQuant as follows. For each block \(i\), we freeze preceding blocks and optimize \(\Theta_1\) and \(\Theta_2\) for block \(i\) by minimizing \(\mathcal L_{\rm block}\). In this approach, all other mechanisms of OmniQuant (such as LWC and LET) remain unchanged. Our approach retains the lightweight nature of PTQ and preserves the per‐block efficient workflow, while enhancing the fidelity of quantized activations in a distributional sense.

Note that our proposed method introduces two additional hyperparameters --namely, the number of projections ($n_{\text{proj}}$) and the weight of the sliced Wasserstein loss (\textbf{sw\_w})-- which should be empirically tuned alongside the hyperparameters of the base quantization method.

A key advantage of the proposed loss lies in its minimal computational overhead during training, attributable to the inherently parallelizable nature of the projection computation. A detailed analysis is presented in Table ~\ref{tab:compute_llama_7b_13b}. Furthermore, as the loss is applied solely during training, it introduces no additional computational cost at inference time.

\section{Results}
\label{s:results}

\subsection{Comparison results based on OmniQuant}
\label{ss:OmniQuant}

Table~\ref{tab:sw_result_llama-2-7b} and Figure~\ref{fig:ppl_comparison} summarize the evaluation results of incorporating the proposed SW loss into OmniQuant measured on the language model LLaMA-2-7B.

Table~\ref{tab:sw_result_llama-2-7b} demonstrates that, across various quantization settings, incorporating the proposed SW loss consistently improves the overall fidelity of the quantized LLaMA-2-7B model, yielding 0.48–2.68 absolute gains in average accuracy over the OmniQuant baseline. This is equivalent to 4.12–20.37\% relative improvements, computed with respect to the gap between the full-precision and baseline OmniQuant models.
The improvement is particularly pronounced in more challenging configurations such as W2A16g128 and W4A4, where the quantization distortion is more severe and the proposed SW loss provides stronger correction.

Figure~\ref{fig:ppl_comparison} shows that similar consistent improvement can be observed in terms of perplexity. Note that lower perplexity indicates that the quantized model predicts the next token distribution more faithfully. 

To ensure that our results for OmniQuant are robust to model sizes, we ran additional experiments on LLaMA-2-13B focusing on the challenging configurations of W2A16g128 and W4A4, identified in the previous experiment.
Table~\ref{tab:sw_result_llama13b} and Figure~\ref{fig:ppl_llama13b_comparison} summarize the comparative results on the proposed method against OmniQuant on LLaMA-2-13B for several configurations. The results consistently show improvements on downstream task accuracy (Table~\ref{tab:sw_result_llama13b}) and language modeling perplexity (Figure~\ref{fig:ppl_llama13b_comparison}) for LLaMA-2-13B.

To ensure that the improvements are not restricted to the case of LLaMA family, we quantify the suitability of the proposed method on the OPT-6.7B language model. Table~\ref{tab:sw_result_opt67b} and Figure~\ref{fig:ppl_opt67b_comparison} summarize the comparative results on the OPT-6.7B model across several quantization configurations. The W4A4 configuration is excluded from the comparison due to gradient explosion during training, which prevented the model from producing meaningful results with or without the SW loss. These results indicate that both downstream task accuracy and perplexity are improved when OmniQuant is augmented with the proposed SW loss function for quantizing OPT-6.7B.

Taken together, the results in Tables \ref{tab:sw_result_llama-2-7b}-\ref{tab:sw_result_opt67b} and Figures \ref{fig:ppl_comparison}--\ref{fig:ppl_opt67b_comparison} demonstrate that the proposed method consistently improves performance across language model sizes and families when incorporated into the OmniQuant method.

To assess the influence of the two hyperparameters that are introduced by our proposed method --namely, the number of projections ($n_{\text{proj}}$) and the weight of the sliced Wasserstein loss (\textbf{sw\_w})--, additional results are presented in Section ~\ref{sec:suppl:additional_results}. To assess the generality of the proposed method beyond OmniQuant, we conduct additional experiments using TesseraQ~\cite{li2024tesseraq}, another state-of-the-art quantization method within the LLMC framework~\cite{gong2024llmc}. The results are presented in Section~\ref{ss:TesseraQ}.

\begin{table*}[htb!]
\centering
\caption{Comparison of Baseline OmniQuant Quantization and the Proposed SW Loss Quantization on LLaMA-2-7B. For each configuration--dataset pair, the highest value is shown in bold.}
\label{tab:sw_result_llama-2-7b}
\renewcommand{\arraystretch}{1.15}

\begin{adjustbox}{max width=\textwidth}
\begin{tabular}{@{}lcccccccccc@{}}
\toprule
\textbf{Config} & \textbf{Calib dataset} & \textbf{$n_{\text{proj}}$} & \textbf{sw\_w} & \textbf{ARC-C} & \textbf{ARC-E} & \textbf{BoolQ} & \textbf{HS} & \textbf{PIQA} & \textbf{Wino} & \textbf{Avg} \\
\midrule
Full Precision & -- & -- & -- & 39.93 & 69.28 & 71.07 & 56.7 & 78.4 & 67.17 & 63.76 \\
\midrule
\multicolumn{11}{c}{\textbf{(a) W2A16g64}} \\
\midrule
OmniQuant & wikitext2 & --  & --   & \textbf{27.39} & \textbf{50.29} & \textbf{63.67} & 42.26 & 67.90 & 56.20 & 51.29 \\
Proposed & wikitext2 & 128 & 0.2 & 27.22 & 49.75 & 63.24 & \textbf{43.90} & \textbf{68.88} & \textbf{59.27} & \textbf{52.04} \\
\midrule
OmniQuant & mixed     & --  & --   & 26.19 & 50.42 & 63.18 & 42.03 & 68.66 & \textbf{57.62} & 51.35 \\
Proposed & mixed     & 128 & 0.2 & \textbf{27.56} & 50.21 & 63.03 & \textbf{43.57} & \textbf{69.75} & 56.83 & \textbf{51.83} \\
\midrule
\multicolumn{11}{c}{\textbf{(b) W2A16g128}} \\
\midrule
OmniQuant  & mixed     & --  & --   & 25.17 & 47.90 & 60.92 & 39.54 & 66.21 & 52.25 & 48.67 \\
Proposed & mixed     & 128 & 0.2 & \textbf{26.88} & \textbf{49.11} & \textbf{63.42} & \textbf{41.91} & \textbf{68.17} & \textbf{56.91} & \textbf{51.07} \\
\midrule
\multicolumn{11}{c}{\textbf{(c) W3A16}} \\
\midrule
OmniQuant & mixed & --   & --   & 35.32 & 62.63 & 64.46 & 51.81 & 74.54 & 64.80 & 58.93 \\
Proposed & mixed & 256  & 0.2 & \textbf{35.58} & \textbf{63.68} & \textbf{64.68} & \textbf{52.50} & \textbf{74.86} & \textbf{66.30} & \textbf{59.60} \\
\midrule
\multicolumn{11}{c}{\textbf{(d) W4A4}} \\
\midrule
OmniQuant & wikitext2 & --   & --   & 23.55 & 41.96 & 60.12 & 36.81 & 61.64 & \textbf{53.35} & 46.24 \\
Proposed & wikitext2 & 256  & 0.1 & \textbf{26.11} & \textbf{48.91} & \textbf{62.05} & \textbf{40.53} & \textbf{63.60} & 52.33 & \textbf{48.92} \\
\bottomrule
\end{tabular}
\end{adjustbox}

\vspace{2mm}
\small
ARC-C = ARC-Challenge, ARC-E = ARC-Easy, HS = HellaSwag, Wino = Winogrande, 
Avg = mean accuracy over ARC-C / ARC-E / BoolQ / HS / PIQA / Wino. $ n_{\text{proj}} $ denotes the number of random 1D projections used in the SW loss, 
and \textbf{sw\_w} refers to the weight assigned to the SW term. 
\end{table*}

\begin{table*}[htb!]
\centering
\caption{Comparison between the OmniQuant Baseline and the Proposed SW Loss Quantization on LLaMA-2-13B. 
For each configuration--dataset pair, the best score per metric is shown in bold.}
\label{tab:sw_result_llama13b}
\renewcommand{\arraystretch}{1.1}
\begin{adjustbox}{max width=\textwidth}
\begin{tabular}{@{}lcccccccccc@{}}
\toprule
\textbf{Config} & \textbf{Calib dataset} & \textbf{$n_{\text{proj}}$} & \textbf{sw\_w} &
\textbf{ARC-C} & \textbf{ARC-E} & \textbf{BoolQ} &
\textbf{HS} & \textbf{PIQA} & \textbf{Wino} & \textbf{Avg} \\
\midrule
Full Precision & -- & -- & -- & 45.56 & 73.27 & 69.02 & 59.70 & 78.73 & 69.53 & 65.97 \\
\midrule
\multicolumn{11}{c}{\textbf{(a) W2A16g128, mixed}} \\
\midrule
OmniQuant & mixed & -- & -- &
29.10 & \textbf{60.06} & 62.57 & 45.72 & 70.29 & 56.27 & 54.00 \\
Proposed & mixed & 256 & 0.1 &
\textbf{29.95} & 58.67 & \textbf{62.75} & \textbf{45.94} & \textbf{71.27} & \textbf{57.06} & \textbf{54.27} \\
\midrule
\multicolumn{11}{c}{\textbf{(b) W4A4, wikitext2}} \\
\midrule
OmniQuant & wikitext2 & -- & -- &
27.65 & 51.68 & 62.63 & 45.51 & \textbf{67.46} & \textbf{53.83} & 51.46 \\
Proposed & wikitext2 & 1024 & 0.05 &
\textbf{29.01} & \textbf{54.67} & \textbf{63.94} & \textbf{46.41} & 66.70 & 53.42 & \textbf{52.36} \\
\bottomrule
\end{tabular}
\end{adjustbox}

\vspace{1mm}
\small
Refer to Table~\ref{tab:sw_result_llama-2-7b} for the legends.
\end{table*}

\begin{table*}[htb!]
\centering
\caption{Comparison between the OmniQuant Baseline and the Proposed SW Loss Quantization on OPT-6.7B. 
For each configuration--dataset pair, the best score per metric is shown in bold.}
\label{tab:sw_result_opt67b}
\renewcommand{\arraystretch}{1.1}
\begin{adjustbox}{max width=\textwidth}
\begin{tabular}{@{}lcccccccccc@{}}
\toprule
\textbf{Calib} & \textbf{Method} & \textbf{$n_{\text{proj}}$} & \textbf{sw\_w} &
\textbf{ARC-C} & \textbf{ARC-E} & \textbf{BoolQ} &
\textbf{HS} & \textbf{PIQA} & \textbf{Wino} & \textbf{Avg} \\
\midrule
Full Precision & -- & -- & -- & 30.63 & 65.70 & 66.09 & 50.48 & 76.28 & 65.27 & 59.08 \\
\midrule
\multicolumn{11}{c}{\textbf{(a) W2A16g128, wikitext2}} \\
\midrule
OmniQuant & wikitext2 & -- & -- &
23.72 & 55.18 & 60.03 & 34.60 & \textbf{67.68} & 53.12 & 49.06 \\
Proposed & wikitext2 & 512 & 0.05 &
\textbf{24.06} & \textbf{55.30} & \textbf{61.16} & \textbf{36.08} & 68.28 & \textbf{54.14} & \textbf{49.84} \\
\midrule
\multicolumn{11}{c}{\textbf{(b) W2A16g128, mixed}} \\
\midrule
OmniQuant & mixed & -- & -- &
23.12 & \textbf{55.18} & 62.11 & \textbf{39.43} & 68.50 & 54.62 & 50.49 \\
Proposed & mixed & 1024 & 0.05 &
\textbf{23.55} & 54.25 & \textbf{62.54} & 39.07 & \textbf{68.61} & \textbf{55.41} & \textbf{50.57} \\
\bottomrule
\end{tabular}
\end{adjustbox}

\vspace{1mm}
\small
Refer to Table~\ref{tab:sw_result_llama-2-7b} for the legends.
\end{table*}

\begin{figure*}[htb!]
    \centering
    \setlength{\tabcolsep}{4pt}
    \renewcommand{\arraystretch}{1.0}
    \begin{tabular}{ccccc}
        \includegraphics[trim={5cm 3cm 1cm 5cm}, clip, width=0.19\textwidth]{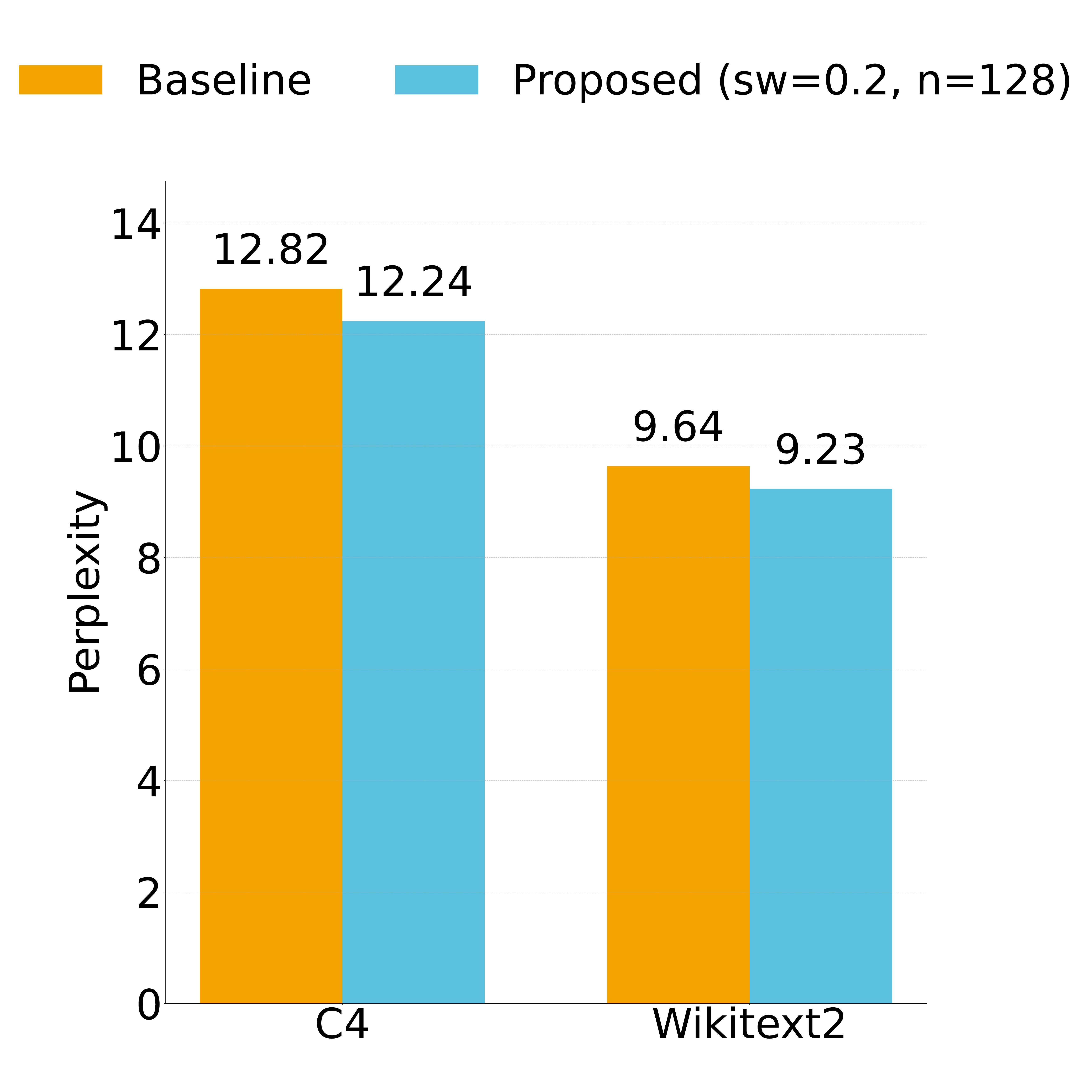} &
        \includegraphics[trim={5cm 3cm 1cm 5cm}, clip, width=0.19\textwidth]{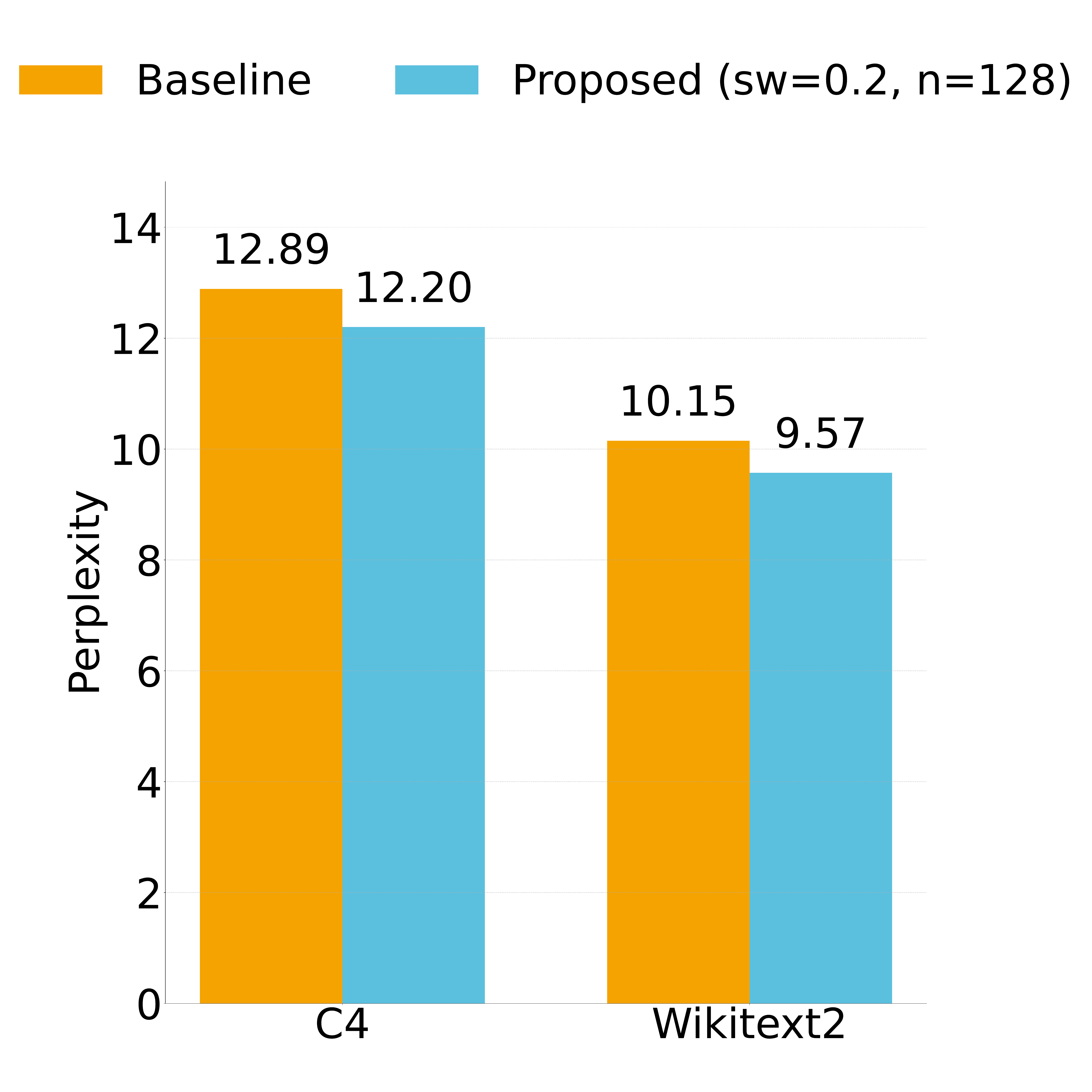} &
        \includegraphics[trim={5cm 3cm 1cm 5cm}, clip, width=0.19\textwidth]{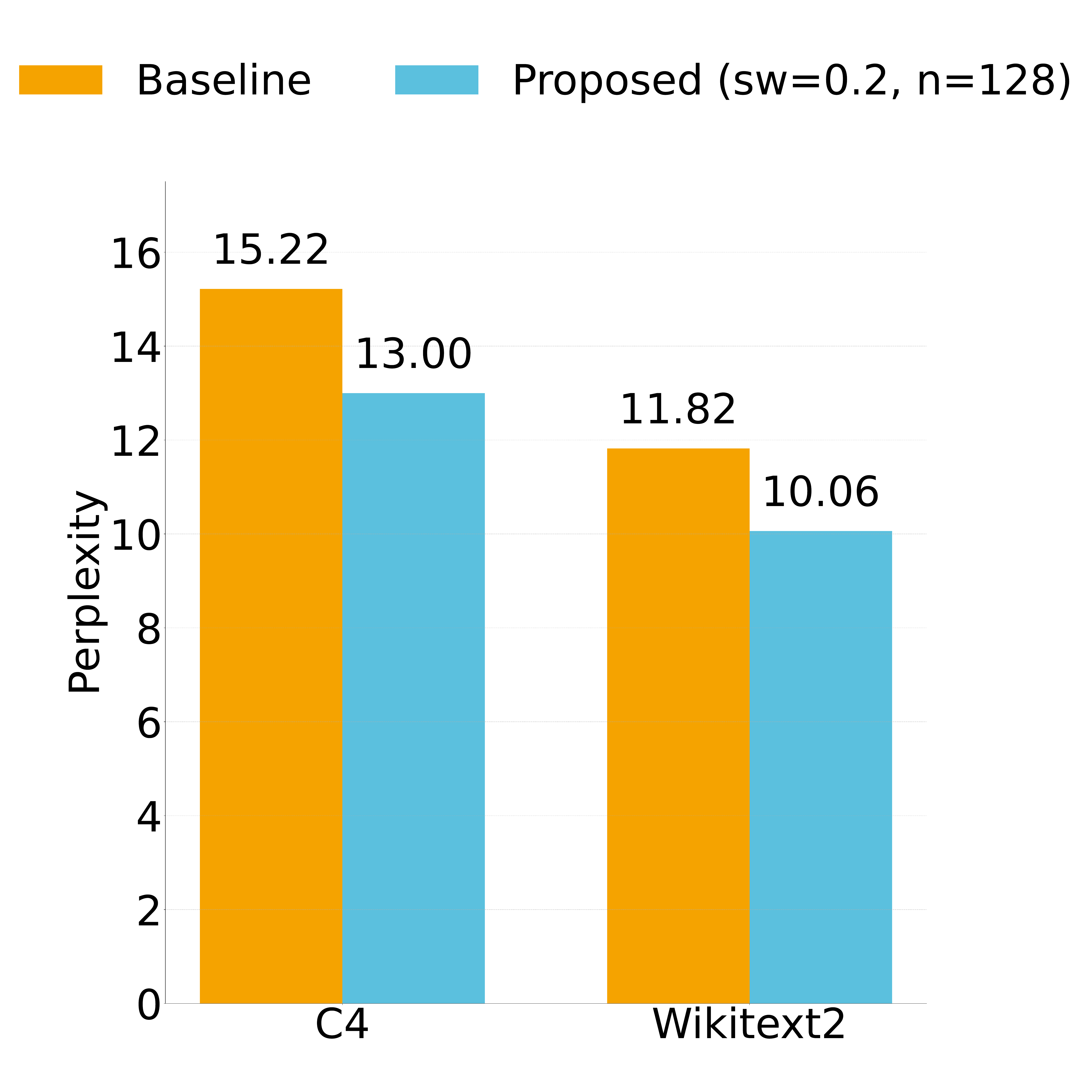} &
        \includegraphics[trim={5cm 3cm 1cm 5cm}, clip, width=0.19\textwidth]{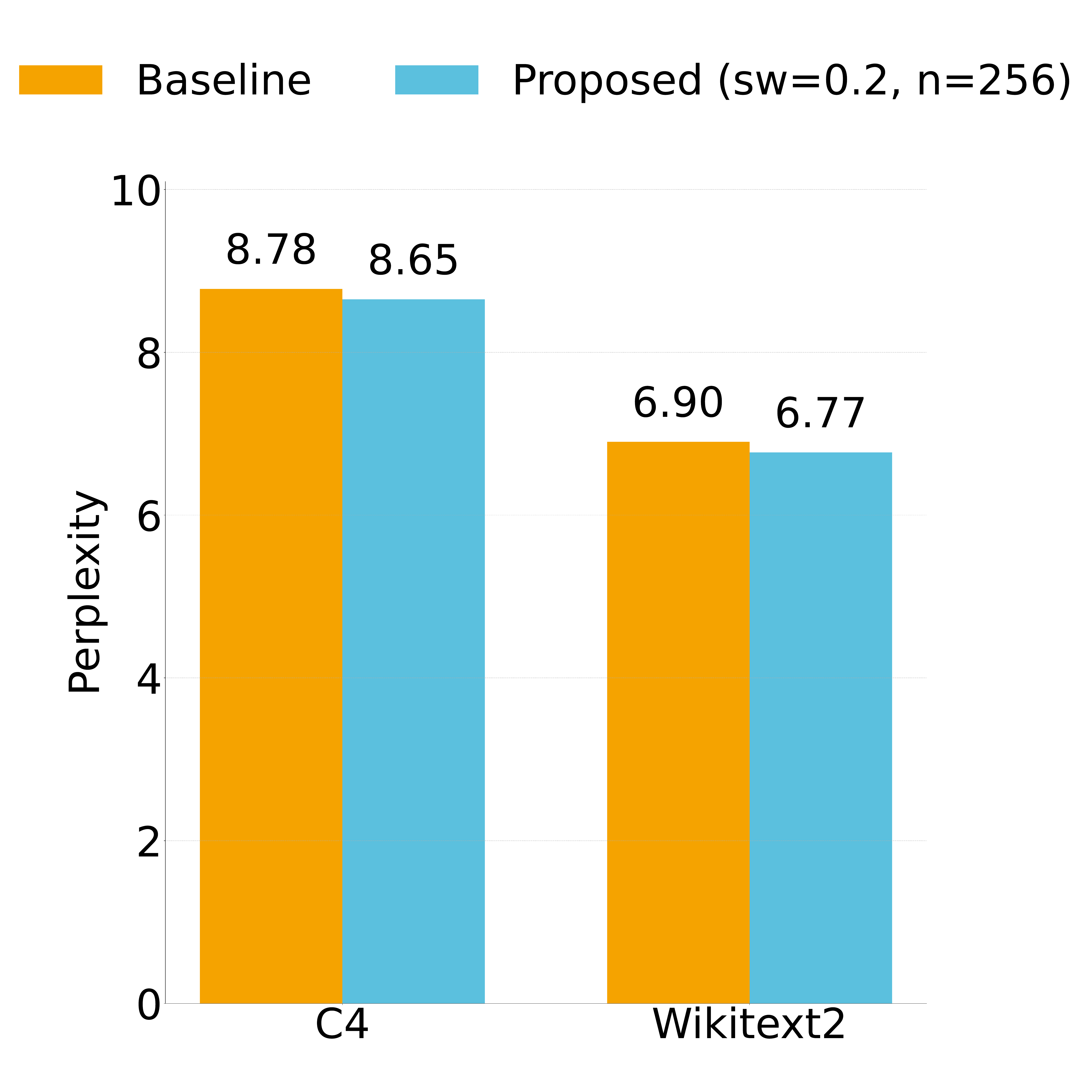} &
        \includegraphics[trim={5cm 3cm 1cm 5cm}, clip, width=0.19\textwidth]{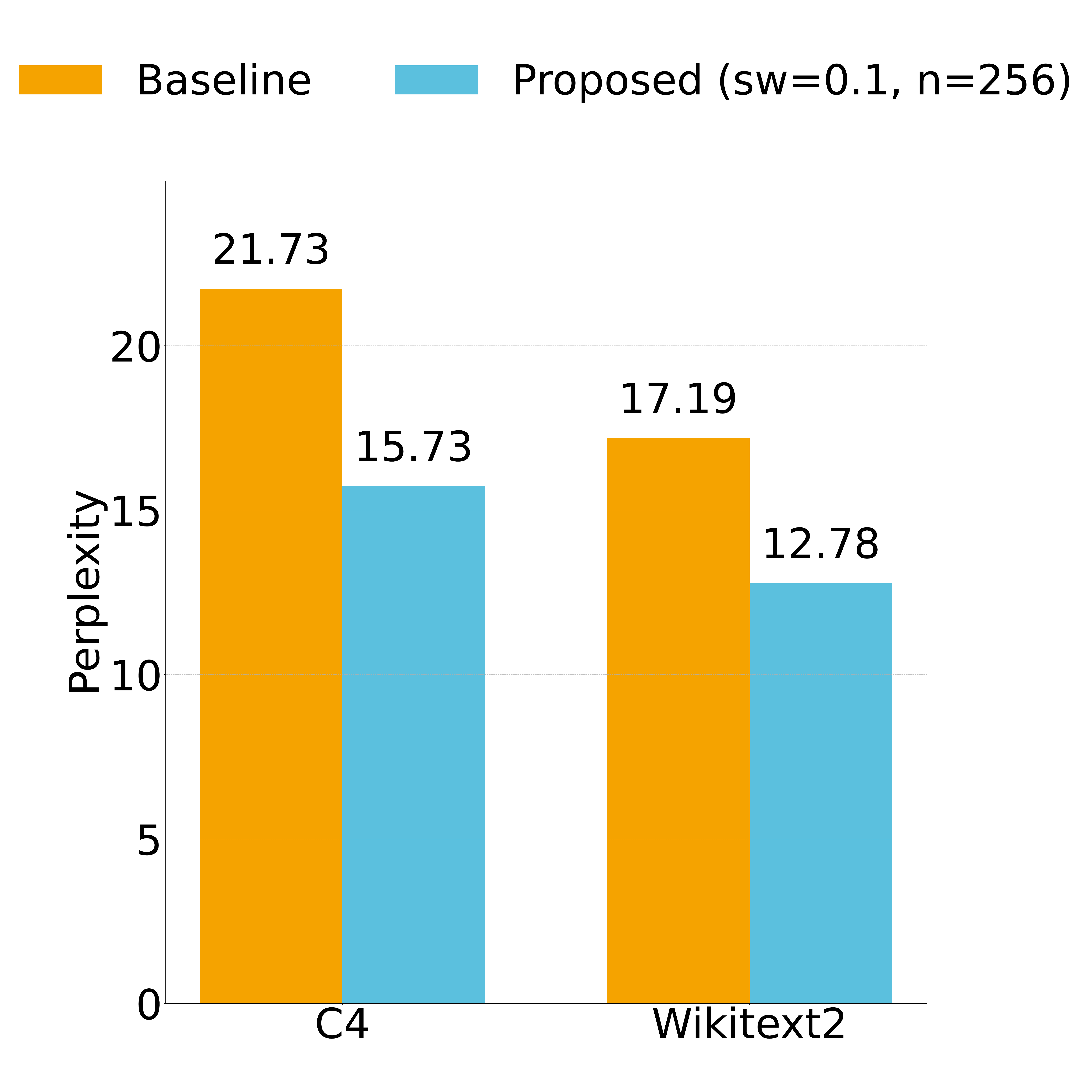} \\
        \small (a) \makecell{W2A16g64\\(wikitext2 calib.)} &
        \small (b) \makecell{W2A16g64\\(mixed calib.)} &
        \small (c) \makecell{W2A16g128\\(mixed calib.)} &
        \small (d) \makecell{W3A16\\(mixed calib.)} &
        \small (e) \makecell{W4A4\\(wikitext2 calib.)}
    \end{tabular}

    \caption{Perplexity comparison between the OmniQuant Baseline and the Proposed SW Loss Quantization on LLaMA-2-7B.}
    \label{fig:ppl_comparison}
\end{figure*}

\begin{figure}[htb!]
    \centering
    \setlength{\tabcolsep}{6pt}
    \renewcommand{\arraystretch}{1.0}
        \includegraphics[trim={0cm 4cm 0cm 4cm}, clip, width=0.22\textwidth]{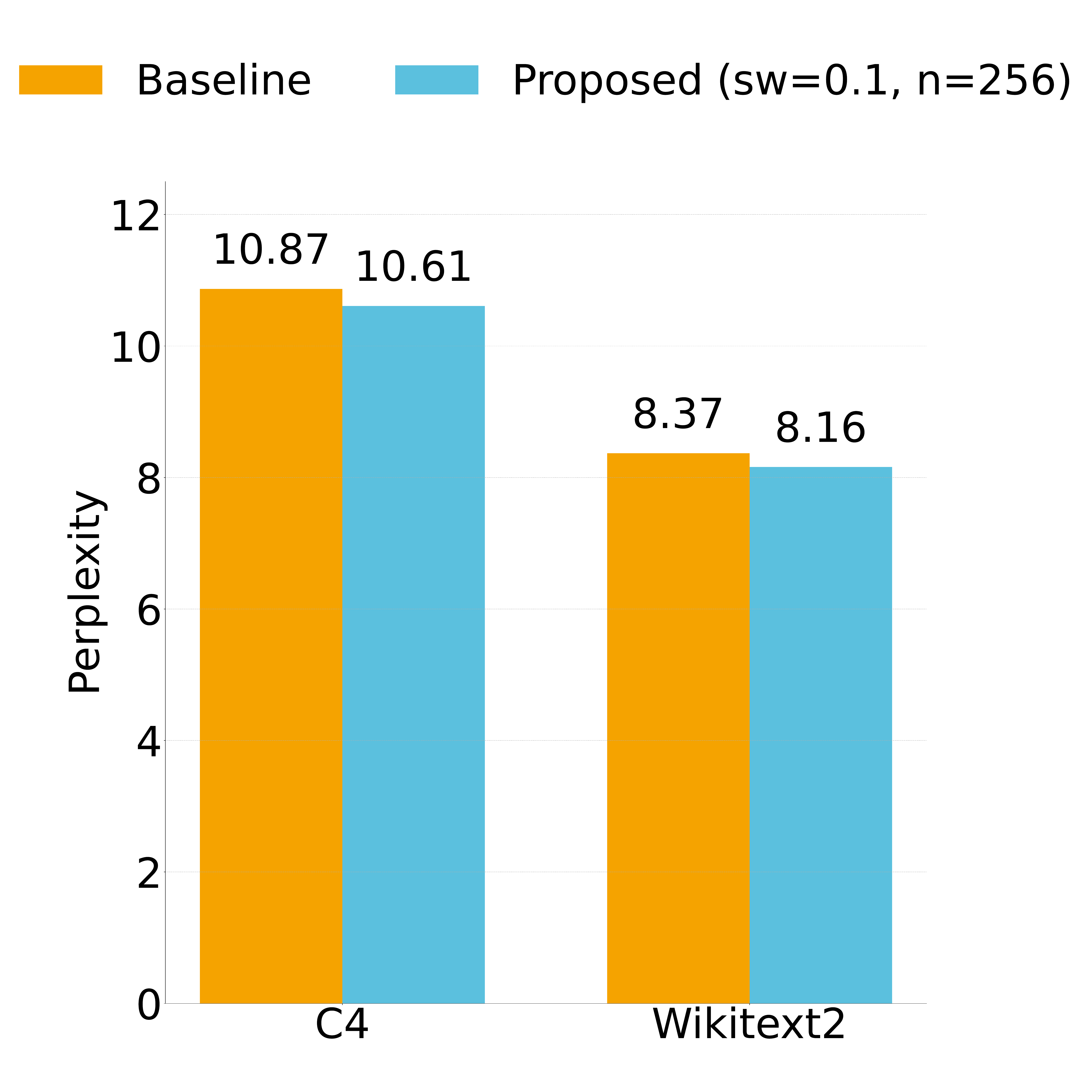} 
        \includegraphics[trim={0cm 4cm 0cm 4cm}, clip, width=0.22\textwidth]{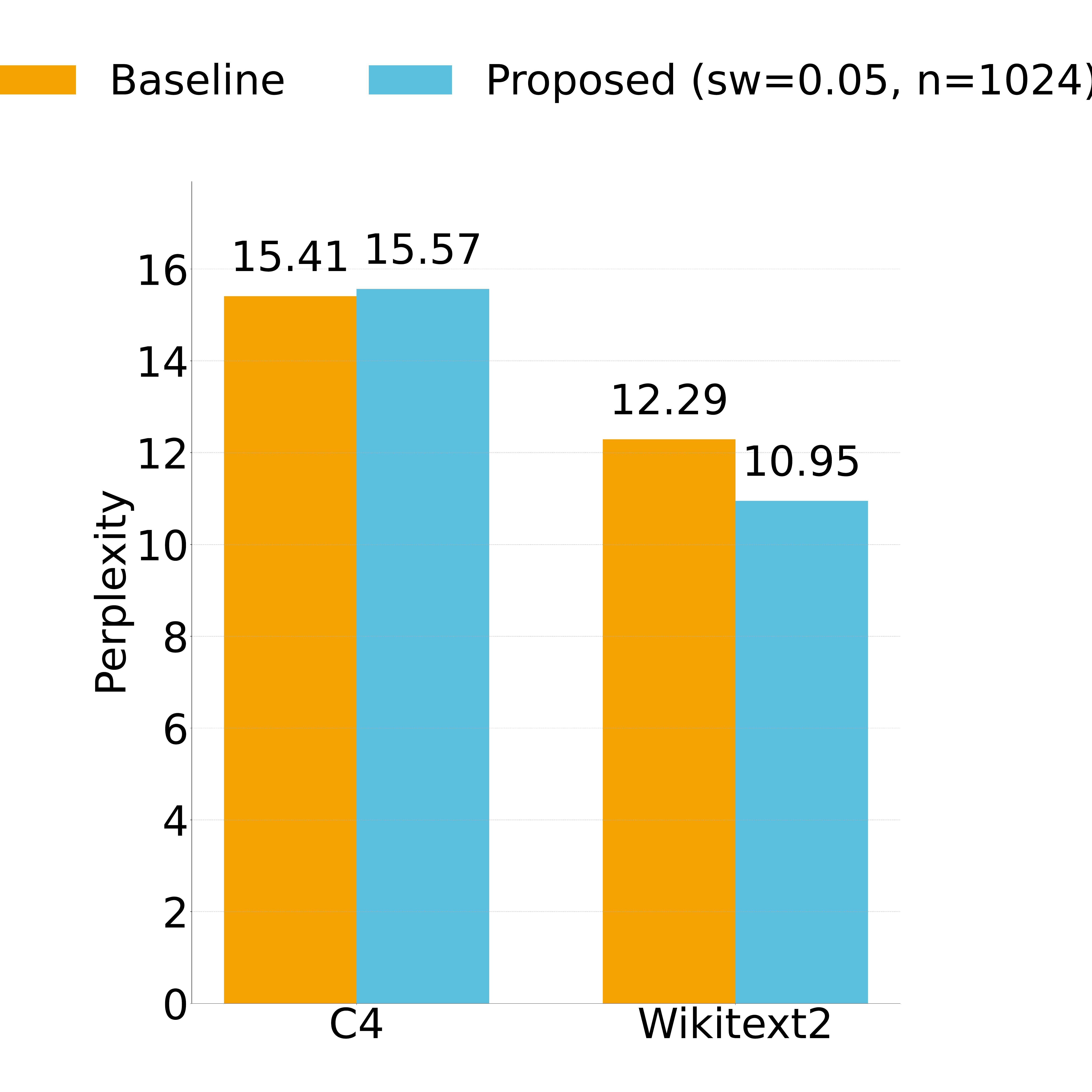}
    \caption{Perplexity comparison between the OmniQuant Baseline and the Proposed SW Loss Quantization on LLaMA-2-13B. Left: LLaMA-2-13B W2A16g128 (mixed calibration). Right: LLaMA-2-13B W4A4 (wikitext2 calibration).
    }
    \label{fig:ppl_llama13b_comparison}
\end{figure}

\begin{figure}[htb!]
    \centering
    \setlength{\tabcolsep}{6pt}
    \renewcommand{\arraystretch}{1.0}
        \includegraphics[trim={0cm 4cm 0cm 4cm}, clip, width=0.22\textwidth]{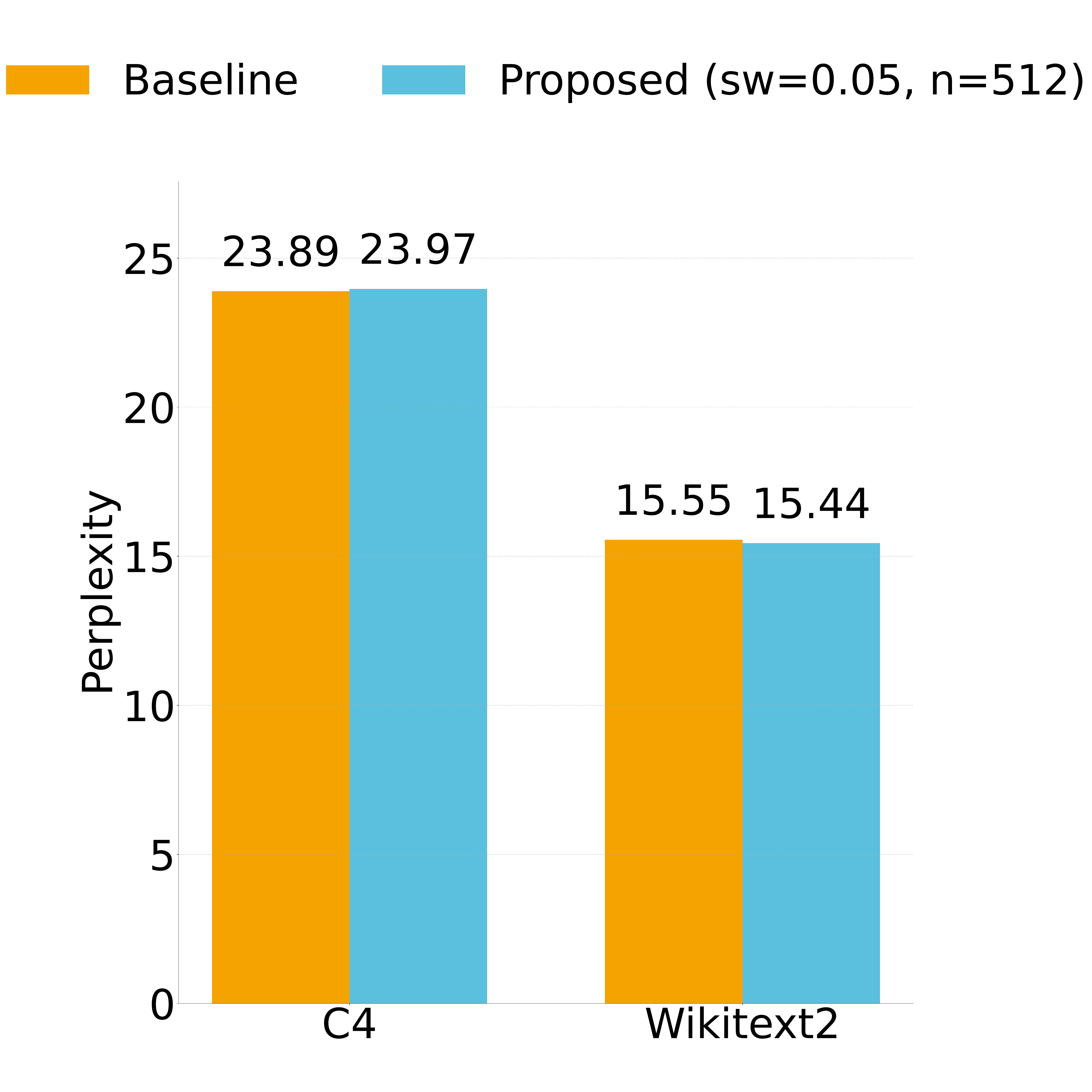} 
        \includegraphics[trim={0cm 4cm 0cm 4cm}, clip, width=0.22\textwidth]{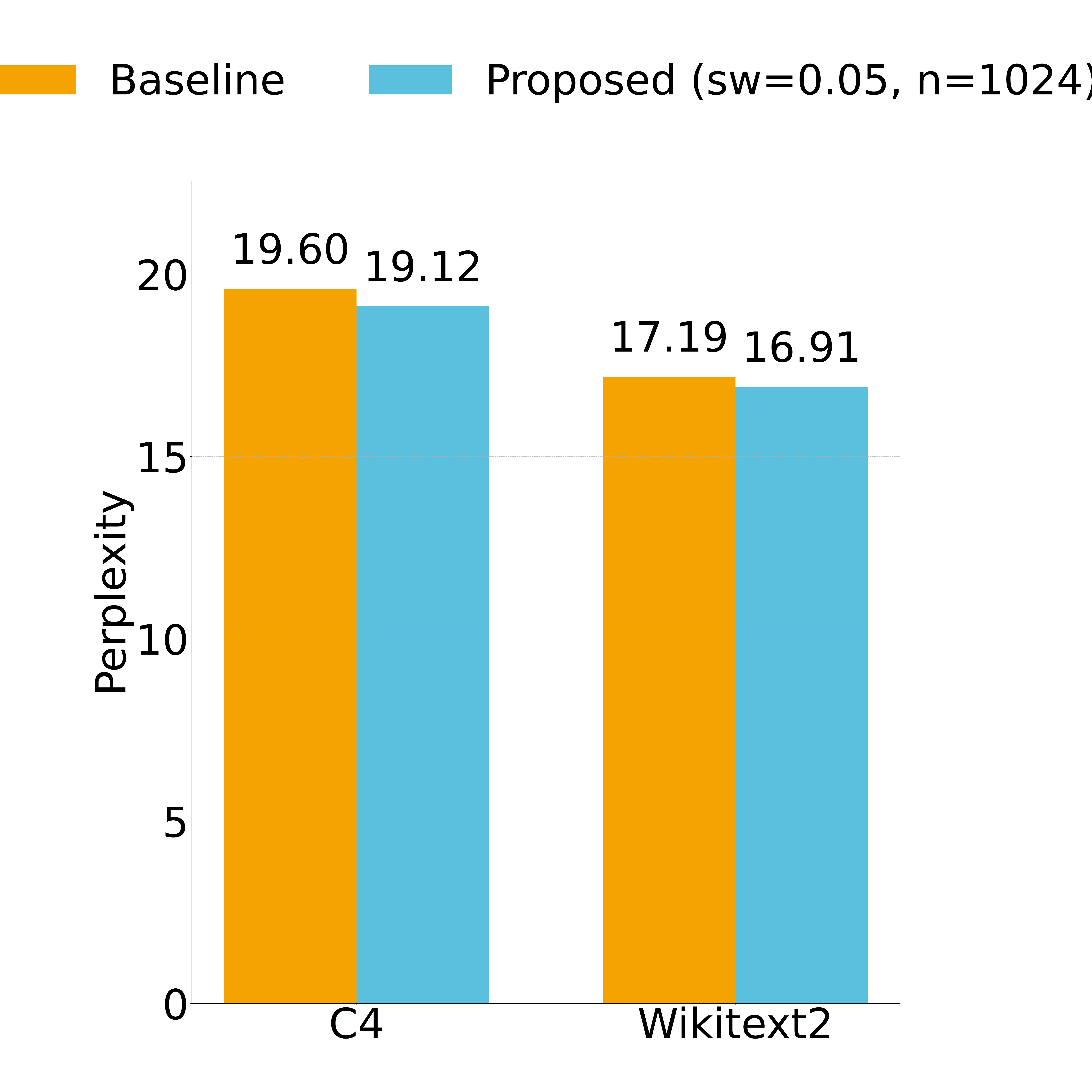} 
    \caption{Perplexity comparison between the OmniQuant Baseline and the Proposed SW Loss Quantization on OPT-6.7B. Left: OPT-6.7B W2A16g128 (wikitext2 calib.). Right: OPT-6.7B W2A16g128 (mixed calib.).
    }
    \label{fig:ppl_opt67b_comparison}
\end{figure}

\subsection{Comparison results based on TesseraQ}
\label{ss:TesseraQ}

\begin{table*}[htb!]
\centering
\caption{Downstream task performance of TesseraQ vs. \textbf{TesseraQ + SW}. 
ARC-C: ARC-Challenge, ARC-E: ARC-Easy, HS: HellaSwag. 
For LLaMA-3.1-8B, BoolQ was not reported in the source table; “—” indicates missing values, and the average is computed over available tasks only.}
\label{tab:sw_on_tesseraq_downstream_all}
\begin{adjustbox}{max width=\textwidth}
\begin{tabular}{@{}lccccccccc@{}}
\toprule
\textbf{Config / Model} & \textbf{$n_{\text{proj}}$} & \textbf{sw\_w} & \textbf{ARC-C} & \textbf{ARC-E} & \textbf{BoolQ} & \textbf{HS} & \textbf{PIQA} & \textbf{Wino} & \textbf{Avg} \\
\midrule
\multicolumn{10}{@{}l}{\textbf{LLaMA-2-7B — W2A16g128}} \\
TesseraQ (repr.)     & --   & --    & 35.70 & 70.10 & \textbf{41.10} & 50.30 & \textbf{75.00} & 59.20 & 55.23 \\
TesseraQ + SW (ours) & 128  & 0.10  & \textbf{36.10} & \textbf{70.55} & 41.07 & \textbf{50.85} & 74.89 & \textbf{59.80} & \textbf{55.54} \\
\midrule
\multicolumn{10}{@{}l}{\textbf{LLaMA-2-7B — W3A16g128}} \\
TesseraQ (repr.)     & --   & --    & 41.35 & 74.35 & 47.10 & 55.25 & \textbf{77.40} & 63.50 & 59.83 \\
TesseraQ + SW (ours) & 256  & 0.10  & \textbf{41.80} & \textbf{74.48} & \textbf{47.45} & \textbf{55.75} & 77.36 & \textbf{63.95} & \textbf{60.13} \\
\midrule
\multicolumn{10}{@{}l}{\textbf{LLaMA-3.1-8B — W2A16g128}} \\
TesseraQ (repr.)\;{\small(5-task)}     & --   & --    & 35.40 & 68.60 & --- & 49.90 & 75.30 & 66.00 & 59.04 \\
TesseraQ + SW (ours)\;{\small(5-task)} & 128  & 0.10  & \textbf{35.55} & \textbf{68.85} & --- & \textbf{50.00} & \textbf{75.55} & \textbf{66.05} & \textbf{59.20} \\
\midrule
\multicolumn{10}{@{}l}{\textbf{LLaMA-3.1-8B — W3A16g128}} \\
TesseraQ (repr.)\;{\small(5-task)}     & --   & --    & 47.10 & 79.10 & --- & 57.50 & 79.00 & 72.60 & 67.06 \\
TesseraQ + SW (ours)\;{\small(5-task)} & 256  & 0.10  & \textbf{47.30} & \textbf{79.35} & --- & \textbf{57.65} & \textbf{79.20} & 72.60 & \textbf{67.22} \\
\bottomrule
\end{tabular}
\end{adjustbox}
\end{table*}

\begin{figure*}
    \centering
    \includegraphics[width=\textwidth,keepaspectratio]{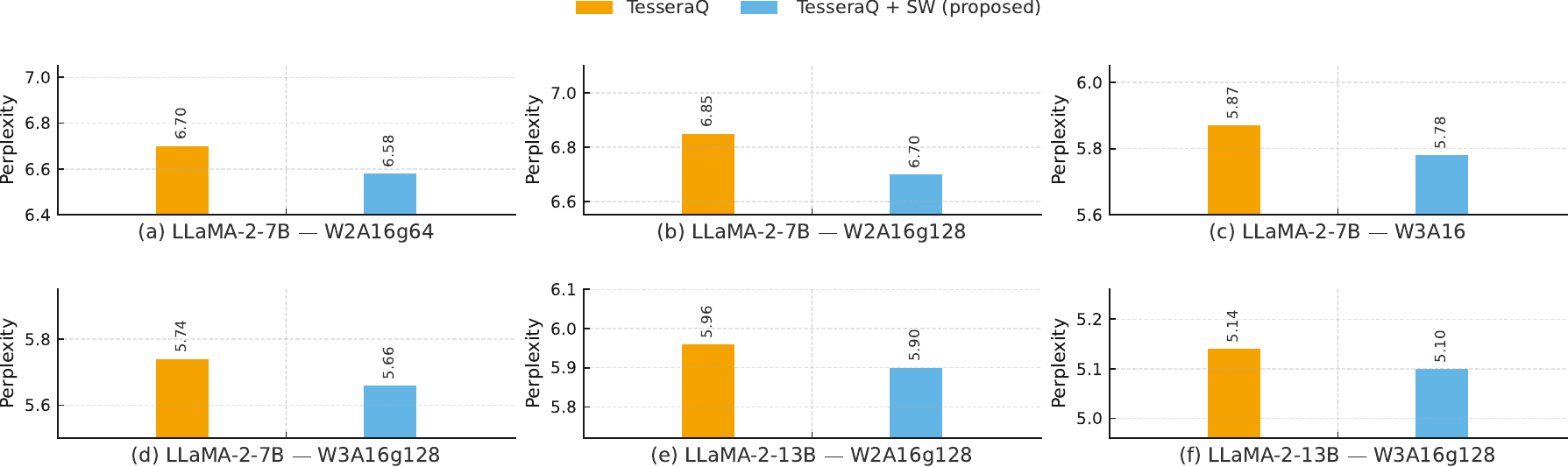}
    \caption{Perplexity comparison for TesseraQ with and without the SW loss on LLaMA-2-7B and LLaMA-2-13B based on WikiText-2 (aggregated panels).}
    \label{fig:tesseraq_ppl_comparison}
\end{figure*}

TesseraQ does not support the OPT-6.7B model. Therefore, we evaluate the proposed SW loss on LLaMA-2-7B, LLaMA-2-13B, and LLaMA-3.1-8B to analyze its applicability when incorporated into TesseraQ. Figure~\ref{fig:tesseraq_ppl_comparison} reports the perplexity results, while Table~\ref{tab:sw_on_tesseraq_downstream_all} summarizes downstream task accuracy.

For the LLaMA-2-7B model, Table~\ref{tab:sw_on_tesseraq_downstream_all} shows that the W2A16g128 configuration achieves the largest absolute improvement: downstream accuracy increases by {0.31 points} (from 55.23 to 55.54), corresponding to a {3.63\% relative gain}. This is smaller than the 1.40\% absolute gain (or 9.28\% relative gain) observed for OmniQuant for the same model and configuration, suggesting that TesseraQ’s block reconstruction and progressive rounding already align quantized features well with the full-precision model, leaving limited room for further improvement by the SW term.

For the other configurations in Table~\ref{tab:sw_on_tesseraq_downstream_all}, similar trends can be observed. 
For {LLaMA-2-7B (W3A16g128)}, the average accuracy increases from 59.83 to 60.13, corresponding to a {0.30 point} improvement ({7.63\% relative gain}). 
For {LLaMA-3.1-8B (W2A16g128)}, the average accuracy rises from 59.04 to 59.20, showing a modest improvement of {0.16 points} ({0.89\% relative gain}). 
Finally, for {LLaMA-3.1-8B (W3A16g128)}, the average improves from 67.06 to 67.22, achieving a {0.16 point} gain ({1.61\% relative gain}).

To further verify the generality of the proposed SW loss, we conducted additional perplexity-based evaluations across a wider range of model scales and quantization configurations.
The results are shown in Figure \ref{fig:tesseraq_ppl_comparison}.
For LLaMA-2-7B, the PPL decreases by approximately {1.39\%–2.19\%} across configurations, while for LLaMA-2-13B, the reduction ranges from {0.78\%–1.00\%}, confirming that the benefit of SW regularization is consistent over various model sizes and bit-width settings.
This difference reflects that larger models, with inherently smoother activations, leave less quantization-induced distortion for the SW term to mitigate.

Overall, the results across Table~\ref{tab:sw_on_tesseraq_downstream_all} and  Figure~\ref{fig:tesseraq_ppl_comparison} confirm that the proposed SW loss provides 
stable, nontrivial benefits for TesseraQ, showing that even a highly optimized baseline 
can still benefit from additional distributional alignment. This observation is 
consistent with prior work that uses KL divergence and other feature reconstruction 
objectives to align full precision and quantized distributions, and that reports 
larger relative gains in aggressive quantization settings
\cite{kim2023squeezellm,gong2024makes}.

\section{Discussion and Future Directions}
\label{sec:discuss}
The experimental results presented both the strengths and the limitations of the proposed approach. Our experiments demonstrated that augmenting OmniQuant’s~\cite{shao2023omniquant} and TesseraQ's~\cite{li2024tesseraq} loss functions with an SW term yields relatively consistent gains in the ultra–low-bit regimes. These gains are recovered accuracy after quantization which can be up to 20.37 percent for OmniQuant and up to 7.63 percent for TesseraQ in relative terms. Across configurations such as W2A16g64, W2A16g128, and the more challenging W4A4 setting, the addition of the SW loss term reduces perplexity and improves downstream task performance, even in the presence of other effective components such as Learnable Weight Clipping (LWC) and Learnable Equivalent Transformation (LET) in the baseline quantization method. Our comprehensive results based on two frontier quantization methods, multiple language models, and multiple model sizes provide support for the hypothesis that element-wise matching alone is insufficient in the ultra-low-bit setting. Therefore, we argue that enforcing distributional alignment at the block level helps preserve higher-order activation structure, thereby making the quantized model more faithful to its full-precision counterpart. This analysis suggests that the prevailing focus on element-wise matching alone is insufficient for ultra-low-bit quantization. Instead, a joint consideration of MSE and SW loss terms, as implemented in our method, can be a compelling solution for the practical deployment of LLMs in a resource-efficient manner with higher after-compression fidelity.

Computationally, incorporating SW introduces a minor overhead due to the required random projections and per-block sorting operations. However, these computations are lightweight and highly parallelizable, resulting in only a marginal increase in time and memory consumption. Please refer to Section ~\ref{sec:suppl:computational_resources} for the additional details.

Overall, these findings are consistent with trends observed in broader quantization research. Studies such as~\cite{dettmers2022calibration} have shown that when quantization becomes more aggressive, for example in 2-bit configurations, with smaller group sizes, or without outlier handling, distribution-aware objectives provide larger benefits than pure mean-squared-error (MSE) calibration. In this context, our method offers a simple plug-and-play augmentation that can be integrated into a wide range of quantization frameworks that involve any form of training.

Future work could explore the integration of our proposed distribution-aware loss function with other frontier quantization frameworks to further validate its generality in both PTQ and QAT contexts. Furthermore, investigating adaptive schemes to balance the MSE and SW loss terms dynamically during training, rather than using a fixed additive weighted average, could yield additional performance gains. Extending the SW loss to more aggressive quantization settings, such as binary or ternary weights, would be a compelling challenge. Finally, our work can be a first step in investigating the suitability and practicality of other distribution-matching loss functions for improving model quantization and compression methods. 

\section*{Code and Reproducibility}
Our proposed method is publicly available on \href{https://github.com/TokuyuSou/SWLoss_LightCompress}{GitHub} at \url{https://github.com/TokuyuSou/SWLoss_LightCompress}.

\FloatBarrier

\section*{Acknowledgements}
This project was supported by grant number SUDSRY4R1P17 from the Data Sciences Institute at the University of Toronto. Authors confirm responsibility for all the content and for their critical and responsible use of AI assistant tools (DeepSeek-V3.2-Exp and ChatGPT-5-Auto) in language and text editing ensuring that the content is accurate.

\clearpage

\appendix
\section*{Supplementary Information}  
\addcontentsline{toc}{section}{Appendix} 
\label{sec:appendix}

\section{Replication of the Baseline Methods}
\label{sec:replication}

To ensure that our results reported based on the quantization framework, OmniQuant, are reliable, we replicated OmniQuant following its official pipeline on large-scale LLMs. Experiments used consistent evaluation on WikiText-2 and C4, covering several common configurations including W3A16 (3-bit weights, 16-bit activations).

\subsection{Replication of OmniQuant}

We replicated the results of OmniQuant~\cite{shao2023omniquant} to establish a reliable baseline to evaluate the effects of our subsequent modifications. Following the setup described in~\cite{shao2023omniquant}, we replicated OmniQuant's quantization on the OPT-6.7B model using the W3A16 configuration (3-bit weights, 16-bit activations), following the official implementation and default hyperparameters provided in~\cite{shao2023omniquant}. The evaluation was conducted on two standard language modeling benchmarks: WikiText2~\cite{merity2016wikitext} and C4~\cite{raffel2020exploring}.

In addition, we replicated the OmniQuant W4A8 quantization setup on the LLaMA-1-7B model, also evaluated on WikiText2 and C4.

As shown in Table~\ref{tab:replication-ppl}, our quantized OPT-6.7B model achieved perplexity values of 11.34 on WikiText2 and 12.42 on C4, while the quantized LLaMA-1-7B model achieved perplexity values of 5.89 on WikiText2 and 7.35 on C4. These results closely match the values reported in the original paper for the same models and bit-width configurations. This consistency demonstrates the reproducibility of OmniQuant across different models and quantization settings, and confirms that our experimental pipeline is well aligned with the original framework.

\begin{table}[htb!]
\caption{Perplexity comparison between our replication and the OmniQuant paper.}
\centering
\resizebox{\linewidth}{!}{
\begin{tabular}{llp{1cm}p{1cm}}
\toprule
{Model, setting} & {Calibration} & {PPL (orig.)} & {PPL (replic.)} \\
\midrule
\multirow{2}{*}{OPT-6.7B, W3A16} &   WikiText2 & 11.65 & \textbf{11.34} \\
                          &   C4        & 12.78 & \textbf{12.42} \\
\midrule
\multirow{2}{*}{LLaMA-1-7B, W4A8} &   WikiText2 & 5.87 & \textbf{5.89} \\
                            &   C4        & 7.34 & \textbf{7.35} \\
\bottomrule
\end{tabular}}
\label{tab:replication-ppl}
\end{table}

\subsection{Replication of TesseraQ}

To ensure our experiments are correct, we independently replicate TesseraQ~\cite{li2024tesseraq}, a post-training quantization (PTQ)
method that optimizes weight rounding via \emph{block reconstruction} and \emph{progressive adaptive rounding} (PAR).
TesseraQ plugs into transformation/clipping PTQ baselines (e.g., AWQ, OmniQuant), further optimizing rounding and
dequantization scales to stabilize reconstruction and improve ultra low-bit performance
(\S4--5 in~\cite{li2024tesseraq}). We focus on LLaMA-2-7B under common settings (W2A16, W3A16, W4A16) and
report both perplexity and five-task zero-shot accuracy.

\paragraph{Perplexity on WikiText-2.}
TesseraQ consistently improves upon AWQ/OmniQuant across bit-widths, with especially large gains at INT2.
In W2A16 (g128), TesseraQ reduces perplexity from 14.65 (AWQ) / 11.06 (OmniQuant) to {6.82};
for W3A16 (g128), from 6.19/6.03 to {5.71}; and for W4A16, from 5.82/5.74 to {5.56}
(see Table~\ref{tab:tesseraq_ppl}). Trends mirror on C4; e.g., W2A16 on 7B improves C4 perplexity from 90.64 to {14.82}
(\cite{li2024tesseraq}, Tbl.~1, Appx.~Tbl.~9). \emph{* indicates initialization from AWQ.}

\begin{table*}[htb!]
\caption{WikiText-2 perplexity (↓) on LLaMA-2-7B; TesseraQ vs.\ strong PTQ baselines~\cite{li2024tesseraq}.}
\centering
\begin{tabular}{lcccc}
\toprule
\textbf{Model} & {Bitwidth} & AWQ & OmniQuant & TesseraQ* \\
\midrule
LLaMA-2-7B & W2A16 (g128) & 14.65 & 11.06 & \textbf{6.82} \\
LLaMA-2-7B & W3A16 (g128) & 6.19 & 6.03 & \textbf{5.71} \\
LLaMA-2-7B & W4A16 & 5.82 & 5.74 & \textbf{5.56} \\
\bottomrule
\end{tabular}
\label{tab:tesseraq_ppl}
\end{table*}

\paragraph{Downstream zero-shot accuracy.}
For weight-only W2A16 (g128) on LLaMA-2-7B, TesseraQ lifts the five-task average from 50.52 (AWQ) and
47.59 (OmniQuant) to {59.27}, approaching the FP16 baseline (64.87) and outperforming SignRound (55.92).
Per-task results are shown in Table~\ref{tab:tesseraq_ds}; average matches the paper (\cite{li2024tesseraq}, Tbl.~2).

\begin{table*}[htb!]
\caption{Zero-shot accuracy (↑) on five tasks for LLaMA-2-7B under W2A16 (g128); data from~\cite{li2024tesseraq}.}
\centering
\begin{tabular}{lcccccc}
\toprule
\textbf{Method (W2A16 g128)} & PIQA & ArcE & ArcC & HellaSwag & WinoG & Avg. \\
\midrule
FP16 & 78.07 & 76.34 & 43.51 & 57.17 & 69.21 & 64.87 \\
GPTQ & 58.21 & 33.75 & 19.79 & 29.60 & 51.30 & 38.53 \\
AWQ & 67.73 & 55.47 & 28.74 & 41.37 & 59.27 & 50.52 \\
OmniQuant & 64.79 & 51.13 & 24.83 & 40.30 & 56.90 & 47.59 \\
SignRound & 72.96 & 65.99 & 32.25 & 47.35 & 61.01 & 55.92 \\
TesseraQ* & \textbf{75.13} & \textbf{70.03} & \textbf{35.83} & \textbf{50.17} & \textbf{65.19} & \textbf{59.27} \\
\bottomrule
\end{tabular}
\label{tab:tesseraq_ds}
\end{table*}

The results indicate that TesseraQ substantially narrows the INT2 gap to FP16 in both perplexity and downstream accuracy, while also yielding consistent gains in W3A16/W4A16.

\section{Other investigations on OmniQuant}

In this section, we present the implementation details and initial results of several preliminary ideas developed based on the OmniQuant framework. These experiments aim to explore potential extensions and modifications to improve quantization performance in large language models.

\subsection{Hierarchical LET: Token and channel-level scaling}

Standard Learnable Equivalent Transformation (LET) in OmniQuant~\cite{shao2023omniquant} applies channel-wise scaling to input activations, adjusting each feature dimension independently to reduce quantization error. However, this approach overlooks token-level variation, particularly the extreme values introduced by rare tokens such as numbers or uncommon words. These token-level outliers can result in highly skewed activation distributions, causing poor quantization fidelity and increased information loss. \\

To address this issue, we propose a hierarchical extension to LET that incorporates both token-level and channel-level normalization. The key idea is to first normalize each token by its own magnitude, effectively aligning the distribution of activations across tokens. This reduces the variance in token-wise activations before applying conventional channel-wise LET. Specifically, we calculate a scaling factor per token \( s_{\text{tok}} = \sqrt{\text{Var}_{\text{channels}}(X)} \), where \( X \) denotes the activation tensor. The activations are then normalized as \( X_1 = X / s_{\text{tok}} \). Following this step, standard LET is applied: \( \tilde{X} = (X_1 - \delta) / s \), where \( \delta \) and \( s \) are learnable channel-wise parameters. \\

This two-stage process helps compress the dynamic range of activations along both the token and channel axes, making them more amenable to quantization. In particular, token-level normalization mitigates the effect of rare-token-induced spikes in activation values, whereas channel-level LET handles distributional differences across dimensions. Together, they enable quantization bins to capture a meaningful signal with reduced distortion. \\

The hierarchical LET method is simple to implement. It only requires computing \( s_{\text{tok}} \) during the calibration phase - prior to actual quantization. These scaling factors can be folded into existing LayerNorm weights, introducing no additional computational cost at inference time. \\

To validate the effectiveness of this method, we conducted preliminary experiments using the OPT-6.7B model under a W4A16 weight-only quantization setup. Our results showed perplexity scores comparable to the original OmniQuant implementation, suggesting that hierarchical LET can maintain performance while offering better control over activation distributions. \\

\subsection{Smarter initialization for LWC}

Learnable Weight Clipping (LWC)\cite{shao2023omniquant} is a core component of OmniQuant, enabling adaptive control of clipping thresholds (\( \gamma \), \( \beta \)) during quantization. Although most existing work focuses on optimizing these parameters during training, our idea investigates whether smarter initialization strategies for \( \gamma \) and \( \beta \) can improve early quantization behavior, particularly at ultra-low bit widths where convergence is fragile and errors are amplified. \\

We explore several heuristics to initialize the clipping parameters before training. These include soft clipping (\( \gamma = \beta = 0.9 \)) to slightly clip outliers while preserving most weight values; aggressive clipping (\( \gamma = \beta = 0.5 \)) to limit the weight range to the central 50\%, increasing resolution near zero at the cost of discarding extreme values; percentile-based initialization (e.g., 95th/5th percentiles) to exclude long-tailed outliers and better capture the true distribution of weights, which are often heavy-tailed in LLMs. \\

Other variants we propose include log-based initialization (for exponentially distributed weights), norm-based initialization (using per-layer norms), noise-aware initialization (based on variance or MAD), and randomized initialization (sampling \( \gamma, \beta \sim U(0.7, 1.0) \)) to assess robustness. \\

Implementation is straightforward: during the calibration phase, we replace the default \( \gamma = \beta = 4 \) with values computed using a single pass over the model's weights. This approach introduces no runtime overhead and can be applied to individual layers or globally. \\

Smarter initialization may improve the quality of the quantification by shrinking the initial dynamic range, reducing the quantification error, stabilizing the gradients for \( \gamma \) and \( \beta \), and accelerating convergence. We hypothesize that this effect is especially impactful at 2-bit or 3-bit precision, where representational flexibility is severely limited. \\

To test this hypothesis, we applied percentile-based initialization on the OPT-6.7B model under W3A8 quantization. Although conceptually promising, this approach resulted in slightly degraded perplexity compared to the default initialization in the original OmniQuant paper. More research is needed to refine the heuristics and understand their interaction with downstream optimization. \\

In general, this line of investigation shifts attention from optimizing clipping thresholds to thoughtfully initializing them, a step that may determine whether training succeeds or diverges, especially in the low-bit quantization regime.

\section{Calibration Loss Functions for Quantization}
\label{appendix:calibration-loss}

A central component of post-training quantization is the \emph{calibration loss function}, which guides the optimization of quantized parameters to approximate the behavior of the full-precision model. While the mean squared error (MSE) between quantized and reference activations is the conventional choice in frameworks such as OmniQuant, alternative objectives may better align with language modeling behavior.

\subsection{Motivation}

The MSE loss focuses on minimizing element-wise activation differences, which does not directly reflect how quantization affects token-level probability distributions. For auto-regressive language models, preserving output distributions is more important than exact feature-level matching. Therefore, we explore distribution-aware calibration losses that explicitly measure divergence between the probability outputs of quantized and full-precision models.

\subsection{Evidence for the potential gains from distribution matching}

To better understand if there are potential performance gains from distribution matching, we performed an initial calibration experiment on the LLaMA-2-7B model using OmniQuant with a W3A16 configuration (3-bit weights, 16-bit activations). Calibration was conducted layer-wise with MSE loss to establish a baseline, and the same setup will later be extended to a KL-based objective.

During this run, the MSE calibration loss decreased from $13.26$ to $12.42$ across the final layer iterations, indicating effective local convergence. However, several iterations produced infinite parameter norms ($\|\mathbf{W}\|_2 \to \infty$), revealing numerical instability under aggressive quantization. We hypothesize that KL-based calibration could stabilize optimization by incorporating probabilistic feedback from the model’s logits rather than relying solely on activation magnitude matching.

\begin{table*}[htb!]
\centering
\caption{Maximum GPU memory and training time comparison between the OmniQuant Baseline and the Proposed SW Loss method on LLaMA-2-7B and LLaMA-2-13B. One representative training configuration is reported per model.}
\label{tab:compute_llama_7b_13b}
\begin{tabular}{@{}ccccc@{}}
\toprule
\textbf{Method} & \textbf{$n_{\text{proj}}$} & \textbf{sw\_w} & \textbf{Max GPU Mem (GB)} & \textbf{Train Time (h)} \\
\midrule
\multicolumn{5}{c}{\textbf{(a) LLaMA-2-7B — W2A16g128, mixed}} \\
\midrule
OmniQuant Baseline & --   & --   & 13.0 & 3.69\\
Proposed (SW)       & 128  & 0.05 & 13.0 & 3.72 \\
\midrule
\multicolumn{5}{c}{\textbf{(b) LLaMA-2-13B — W2A16g128, mixed}} \\
\midrule
OmniQuant Baseline & --   & --   & 17.6 & 4.27 \\
Proposed (SW)      & 256  & 0.1  & 17.6 & 4.29 \\
\bottomrule
\end{tabular}
\vspace{1mm}
\end{table*}

\begin{figure*}[htb!]
    \centering
    \setlength{\tabcolsep}{6pt}
    \renewcommand{\arraystretch}{1.0}
    \begin{tabular}{cc}
        \includegraphics[width=0.44\textwidth]{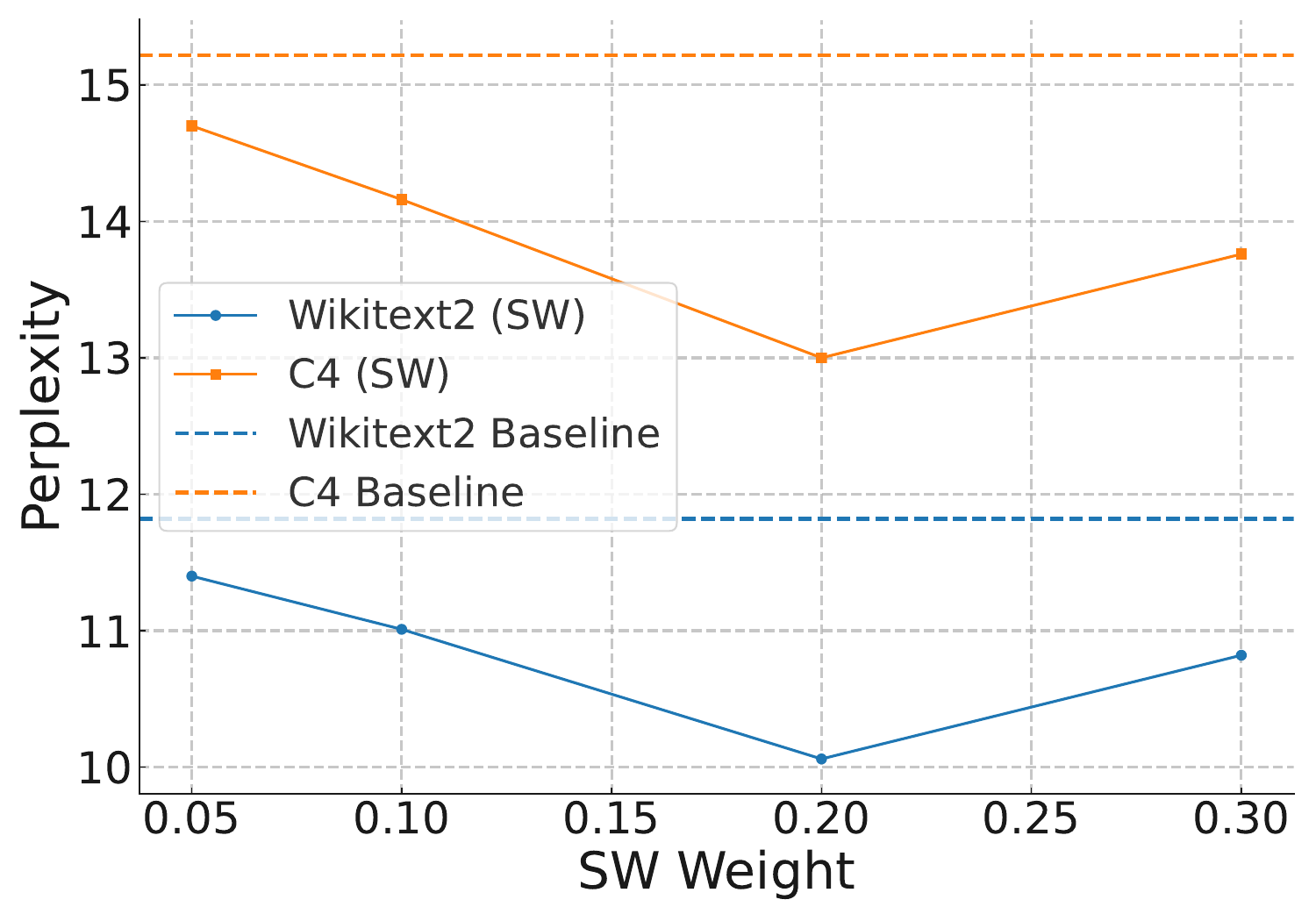} &
        \includegraphics[width=0.44\textwidth]{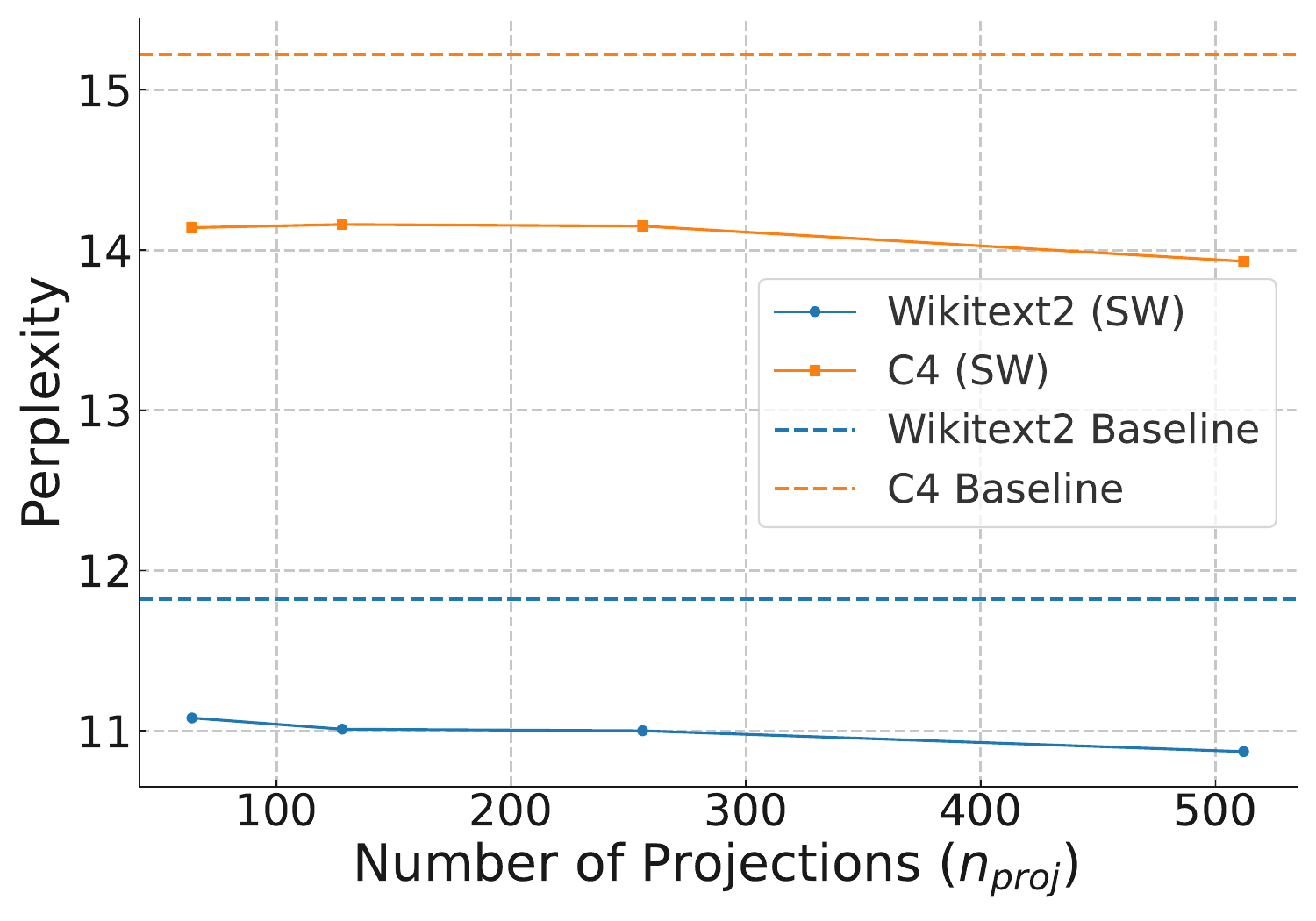} \\
        \small (a) PPL vs.\ $w_{\text{SW}}$ at $n_{\text{proj}}{=}128$ &
        \small (b) PPL vs.\ $n_{\text{proj}}$ at $w_{\text{SW}}{=}0.1$
    \end{tabular}
    \caption{Impact of the SW hyperparameters on quantization performance of LLaMA-2-7B under the W2A16g128 configuration with \textbf{mixed} calibration. 
    Figure~(a) examines how varying the SW loss weight ($w_{\text{SW}}$) affects the perplexity when $n_{\text{proj}}=128$, while Figure~(b) analyzes the influence of the number of projection samples ($n_{\text{proj}}$) when $w_{\text{SW}}=0.1$. 
    The dashed lines indicate the OmniQuant baseline, showing that the proposed SW-loss quantization achieves lower perplexity with minimal sensitivity to these hyperparameters.}
    \label{fig:ppl_llama2_7b_swweight_nproj}
\end{figure*}

\FloatBarrier


\subsection{Outlook}

Although this KL calibration has not yet been fully implemented, the motivation and preliminary results suggest it as a promising direction. Future work will compare MSE and KL objectives quantitatively in terms of perplexity, stability, and distribution alignment, and investigate hybrid formulations combining both losses:
\[
\mathcal{L}_{\text{hybrid}} = \alpha\,\mathcal{L}_{\mathrm{MSE}} 
+ (1-\alpha)\,\mathcal{L}_{\mathrm{KL}} ,
\]
where $\alpha \in [0,1]$ balances structural and distributional fidelity.

This exploration laid the groundwork for our main experiments to improve ultra-low-bit quantization stability through alternative distribution-matching loss functions.

\section{Computation and Resource Requirements}
\label{sec:suppl:computational_resources}
As shown in Table~\ref{tab:compute_llama_7b_13b}, the proposed SW loss quantization introduces only a marginal increase in GPU memory usage and training time compared to the OmniQuant baseline. While increasing $n_{\text{proj}}$ slightly prolongs the training process, the overhead remains within a few percent even at higher settings such as $n_{\text{proj}} = 1024$ or $2048$. These results demonstrate that the additional SW-based optimization can be seamlessly integrated into existing quantization workflows with minimal computational and memory overhead.

\section{Additional Results}
Figure~\ref{fig:ppl_llama2_7b_swweight_nproj} illustrates how the proposed SW loss quantization behaves under different hyperparameter settings. Specifically, it visualizes the effect of varying the SW loss weight $w_{\text{SW}}$ and the number of projection samples $n_{\text{proj}}$ on the perplexity measured on the Wikitext2 and C4 datasets.
Based on the results in Figure~\ref{fig:ppl_llama2_7b_swweight_nproj}, we observe that the balance between the MSE loss and the Sliced-Wasserstein loss, controlled by $w_{\text{SW}}$, plays an important role in determining quantization quality. An appropriate intermediate value of $w_{\text{SW}}$ tends to yield the best performance, indicating the existence of an optimal trade-off between reconstruction fidelity and distributional alignment. Increasing the number of projection samples $n_{\text{proj}}$ generally improves accuracy and reduces perplexity, although the improvement becomes marginal beyond a certain point. While these trends are generally observed under different configurations, the optimal combination of $w_{\text{SW}}$ and $n_{\text{proj}}$ may vary depending on the model architecture and quantization configuration (e.g., bit width), suggesting that a dedicated hyperparameter search is required for each setting.
\label{sec:suppl:additional_results}

\end{document}